\def\year{2020}\relax
\documentclass[letterpaper]{article} % DO NOT CHANGE THIS
\usepackage{aaai20}  % DO NOT CHANGE THIS
\usepackage{times}  % DO NOT CHANGE THIS
\usepackage{helvet} % DO NOT CHANGE THIS
\usepackage{courier}  % DO NOT CHANGE THIS
\usepackage[hyphens]{url}  % DO NOT CHANGE THIS
\usepackage{graphicx} % DO NOT CHANGE THIS
\usepackage{graphicx}  % DO NOT CHANGE THIS
\usepackage{xcolor}
\usepackage{amsmath}
\usepackage{amsthm}
\usepackage{amssymb}
\theoremstyle{definition}
\newtheorem{definition}{Definition}
\newtheorem{theorem}{Theorem}
\newcommand{\del}[1]{}
\newcommand\norm[1]{\lVert#1\rVert}
\newcommand{\mtx}[1]{\ensuremath{\mathbf{#1}}}
\usepackage[linesnumbered, ruled, vlined]{algorithm2e}

\SetCommentSty{mycommfont}

\usepackage[caption=false]{subfig}
\usepackage{stmaryrd}
\usepackage{multirow}
\newcommand{\citet}[1]{\citeauthor{#1} \shortcite{#1}}

\newcommand{\mc}[1]{\mathcal{#1}}

\title{Practical Federated Gradient Boosting Decision Trees}
\author{
Qinbin Li,\textsuperscript{\rm 1}
Zeyi Wen,\textsuperscript{\rm 2}
Bingsheng He\textsuperscript{\rm 1}\\
\textsuperscript{\rm 1}National University of Singapore\\
\textsuperscript{\rm 2}The University of Western Australia\\
\{qinbin, hebs\}@comp.nus.edu.sg, zeyi.wen@uwa.edu.au
}
\begin{document}

\maketitle

\begin{abstract}
Gradient Boosting Decision Trees (GBDTs) have become very successful in recent years, with many awards in machine learning and data mining competitions. There have been several recent studies on how to train GBDTs in the federated learning setting. In this paper, we focus on horizontal federated learning, where data samples with the same features are distributed among multiple parties. However, existing studies are not efficient or effective enough for practical use. They suffer either from the inefficiency due to the usage of costly data transformations such as secret sharing and homomorphic encryption, or from the low model accuracy due to differential privacy designs. In this paper, we study a practical federated environment with relaxed privacy constraints. In this environment, a dishonest party might obtain some information about the other parties' data, but it is still impossible for the dishonest party to derive the actual raw data of other parties. Specifically, each party boosts a number of trees by exploiting similarity information based on locality-sensitive hashing. We prove that our framework is secure without exposing the original record to other parties, while the computation overhead in the training process is kept low. Our experimental studies show that, compared with normal training with the local data of each party, our approach can significantly improve the predictive accuracy, and achieve comparable accuracy to the original GBDT with the data from all parties.
\end{abstract}

\section{Introduction}
Federated learning (FL)~\cite{McMahan2016FederatedLO,mirhoseini2016cryptoml,shi2017distributed,Yang:2019:FML:3306498.3298981,pmlr-v97-mohri19a,li2019federated} has become a hot research area in machine learning. Federated learning addresses the privacy and security issues of model training in multiple parties. In reality, data are dispersed over different areas. For example, people tend to go to nearby hospitals, and the patient records in different hospitals are isolated. Ideally, hospitals may benefit more if they can collaborate with each other to train a model with the joint data. However, due to the increasing concerns and more regulations/policies on data privacy, organizations are not willing to share their own raw data records. Also, according to a recent survey~\cite{Yang:2019:FML:3306498.3298981}, federated learning can be broadly categorized into \emph{horizontal} federated learning, \emph{vertical} federated learning and federated transfer learning. Much research efforts have been devoted to developing new learning algorithms in the setting of vertical or horizontal federated learning~\cite{smith2017federated,takabi2016privacy,liu2018secure,pmlr-v97-yurochkin19a}. We refer readers for more recent surveys for details~\cite{Yang:2019:FML:3306498.3298981,li2019federated}.

On the other hand, Gradient Boosting Decision Trees (GBDTs) have become very successful in recent years by winning many awards in machine learning and data mining competitions~\cite{chen2016xgboost} as well as their effectiveness in many applications~\cite{richardson2007predicting,kim2009improving,burges2010ranknet,li2019privacy}. There have been several recent studies on how to train GBDTs in the federated learning setting~\cite{cheng2019secureboost,liu2019boosting,zhao2018inprivate}. For example, SecureBoost~\cite{cheng2019secureboost} developed vertical learning with GBDTs. In contrast, this study focuses on horizontal learning for GDBTs, where data samples with the same features are distributed among multiple parties.

There have been several studies of GDBT training in the setting of horizontal learning~\cite{liu2019boosting,zhao2018inprivate}. However, those approaches are not effective or efficient enough for practical use. 

\emph{Model accuracy:}  The learned model may not have a good predictive accuracy. A recent study adopted differential privacy to aggregate distributed regression trees~\cite{zhao2018inprivate}. This approach boosts each tree only with the local data, which does not utilize the information of data from other parties. As we will show in the experiments, the model accuracy is much lower than our proposed approach.

\emph{Efficiency:} The approach~\cite{liu2019boosting} has a prohibitively time-consuming learning process since it adopts complex cryptographic methods to encrypt the data from multiple parties. Due to a lot of extra cryptographic calculations, the approach brings prohibitively high overhead in the training process. Moreover, since GBDTs have to traverse the feature values to find the best split value, there is a huge number of comparison operations even in the building of a single node. 

%It is prohibitively time-consuming to adopt encryption each time performing an operation. 

Considering the previous approaches' limitations on efficiency and model accuracy, this study utilizes a more practical privacy model as a tradeoff between privacy and efficiency/model accuracy~\cite{du2004privacy,liu2018secure}. In this environment, a dishonest party might obtain some information about the other parties' data, but \emph{it is still impossible for the dishonest party to derive the actual raw data of other parties}. Compared to differential privacy or secret sharing, this privacy model is weaker, but enables new opportunities for designing much more efficient and effective GBDTs. 

Specifically, we propose a novel and practical federated learning framework for GDBTs (named SimFL). The basic idea is that instead of encryption on the feature values, we make use of the similarity between data of different parties in the training while protecting the raw data. First, we propose the use of locality-sensitive hashing (LSH) in the context of federated learning. We adopt LSH to collect similarity information without exposing the raw data.  Second, we design a new approach called Weighted Gradient Boosting (WGB), which can build the decision trees by exploiting the similarity information with bounded errors. Our analysis show that SimFL satisfies the privacy model~\cite{du2004privacy,liu2018secure}. The experimental results show that SimFL shows a good accuracy, while the training is fast for practical uses. 

\section{Preliminaries}

\paragraph{Locality-Sensitive Hashing (LSH)}
LSH was first introduced by~\citet{gionis1999similarity} for approximate nearest neighbor search. The main idea of LSH is to select a hashing function such that (1) the hash values of two neighbor points are equal with a high probability and (2) the hash values of two non-neighbor points are \emph{not} equal with a high probability. A good property of LSH is that there are infinite input data for the same hash value. Thus, LSH has been used to protect user privacy in applications such as keyword searching~\cite{wang2014privacy} and recommendation systems~\cite{qi2017distributed}. 

The previous study~\cite{datar2004locality} proposed the $p$-stable LSH family, which has been widely used. The hash function $\mathcal{F}_{\mathbf{a},b}$ is formulated as $\mathcal{F}_{\mathbf{a},b}(\mathbf{v})=\lfloor\frac{\mathbf{a}\cdot\mathbf{v}+b}{r}\rfloor$, where $\mathbf{a}$ is a $d$-dimensional vector with entries chosen independently from a $p$-stable distribution~\cite{zolotarev1986one}; $b$ is a real number chosen uniformly from the range $[0,r]$; $r$ is a positive real number which represents the window size.

\paragraph{Gradient Boosting Decision Trees (GBDTs)}
The GBDT is an ensemble model which trains a sequence of decision trees. Formally, given a loss function $l$ and a dataset with $n$ instances and $d$ features $\mathcal{D}=\{(\mathbf{x}_i, y_i)\} (|\mathcal{D}|=n, \mathbf{x}_i\in \mathbb{R}^d, y_i\in \mathbb{R})$, 
GBDT minimizes the following objective function~\cite{chen2016xgboost}.
\begin{equation}
    \mathcal{\tilde{L}} = \sum_il(\hat{y}_i,y_i)+\sum_k \Omega(f_k) 
\end{equation}
where $\Omega(f)= \gamma T_l + \frac{1}{2}\lambda \norm{w}^2$ is a regularization term to penalize the complexity of the model. Here $\gamma$ and $\lambda$ are hyper-parameters, $T_l$ is the number of leaves and $w$ is the leaf weight. Each $f_k$ corresponds to a decision tree. 
Training the model in an additive manner, GBDT minimizes the following objective function at the $t$-th iteration.
\begin{equation}
    \mathcal{\tilde{L}}^{(t)} = \sum_{i=1}^{n} [g_i f_t(\mathbf{x}_i)+\frac{1}{2}h_i f_t^2(\mathbf{x}_i)]+\Omega(f_t)
\label{eq:xg_loss}
\end{equation}
where $g_i = \partial_{\hat{y}^{(t-1)}}l(y_i,\hat{y}^{(t-1)})$ and $h_i = \partial_{\hat{y}^{(t-1)}}^2l(y_i,\hat{y}^{(t-1)})$ are first and second order gradient statistics on the loss function. The decision tree is built from the root until reaching the restrictions such as the maximum depth. Assume $I_L$ and $I_R$ are the instance sets of left and right nodes after the split. Letting $I = I_L \cup I_R$, the gain of the split is given by
\begin{small}
\begin{equation}
    \mathcal{L}_{split} = \frac{1}{2}[\frac{(\sum_{i\in I_L}g_i)^2}{\sum_{i\in I_L}h_i+\lambda}+\frac{(\sum_{i\in I_R}g_i)^2}{\sum_{i\in I_R}h_i+\lambda}-
    \frac{(\sum_{i\in I}g_i)^2}{\sum_{i\in I}h_i+\lambda}]-\gamma
\label{eq:gain}
\end{equation}
\end{small}

GBDT traverses all the feature values to find the split that maximizes the gain.

\section{Problem Statement}
% while each party $P_i$ cannot tell whether an instance $x_q$ belongs to $I_j$ $(j\neq i)$.
This paper focuses on the application scenarios of horizontal federated learning. Multiple parties have their own data which have the same set of features. Due to data privacy requirements, they are not willing to share their private data with other parties. However, all parties want to exploit collaborations and benefits from a more accurate model that can be built from the joint data from all parties. Thus, \emph{the necessary incentive for this collaboration is that federated learning should generate a much better learned model than the one generated from the local data of each party alone.} In other words, (much) better model accuracy is a pre-condition for such collaborations in horizontal federated learning. We can find such scenarios in various applications such as banks and healthcares~\cite{Yang:2019:FML:3306498.3298981}.

Specifically, we assume that there are $M$ parties, and each party is denoted by $P_i$ $(i\in [1,M])$. We use $I_m = \{(\mathbf{x}_i^m, y_i^m)\}$ $(|I_m|=N_m, \mathbf{x}_i^m\in \mathbb{R}^d, y_i^m\in \mathbb{R})$ to denote the instance set of $P_m$. For simplicity, the instances have global IDs that are unique identifiers among parties (i.e., given two different instances $\mathbf{x}_i^m$ and $\mathbf{x}_j^n$, we have $i \neq j$).

{\bf Privacy model.} The previous study~\cite{du2004privacy} proposed a 2-party security model, which is also adopted in the previous studies (e.g.,~\cite{liu2018secure}). We extend the model for multiple parties, and we get the following privacy definition.

\begin{definition}[Privacy Model]
Suppose all parties are \textit{honest-but-curious}. For a protocol $C$ performing $(O_1, O_2, ..., O_M) = C(I_1, I_2, ..., I_M)$, where $O_1, O_2, ..., O_M$ are parties $P_1, P_2, ..., P_M$'s output and $I_1, I_2, ..., I_M$ are their inputs, $C$ is secure against $P_1$ if there exists infinite number of tuples $(I_2', ..., I_M', O_2', ..., O_M')$\del{ $(j \in [1,M],j\neq k)$} such that $(O_1, O_2', ..., O_M') = C(I_1, I_2', ..., I_M')$. 
\label{def:security}
\end{definition}

Compared with the security definition in secure multi-party computation~\cite{yao1982protocols}, the privacy model in Definition~\ref{def:security} is weaker in the privacy level, as also discussed in the previous study~\cite{du2004privacy}. It does not handle the potential risks such as inference attacks. For example, it can happen that all the possible inputs for a certain output are close to each other, enough information about the input data may be disclosed (even though the exact information for the input is unknown). However, how likely those attacks can happen and its impact in real applications are still to be determined. Thus, like the previous studies~\cite{du2004privacy,liu2018secure}, we view this model as a heuristics model for privacy protection. More importantly, with such a heuristic model, it is possible that we develop practical federated learning models that are much more efficient and have much higher accuracy than previous approaches.

{\bf Problem definition.} The objective is to build an efficient and effective GBDT model under the privacy model in Definition~\ref{def:security} over the instance set $I = \bigcup_{i=1}^MI_i$ $(|I|=N)$.

\section{The SimFL Framework}
In this section, we introduce our framework, Similarity-based Federated Learning (SimFL), which enables the training of GBDTs in a horizontally federated setting.

\begin{figure}[t]
% \captionsetup[subfloat]{farskip=2pt,captionskip=1pt}
\centering
\subfloat[The preprocessing stage]{\includegraphics[width=0.95\columnwidth]{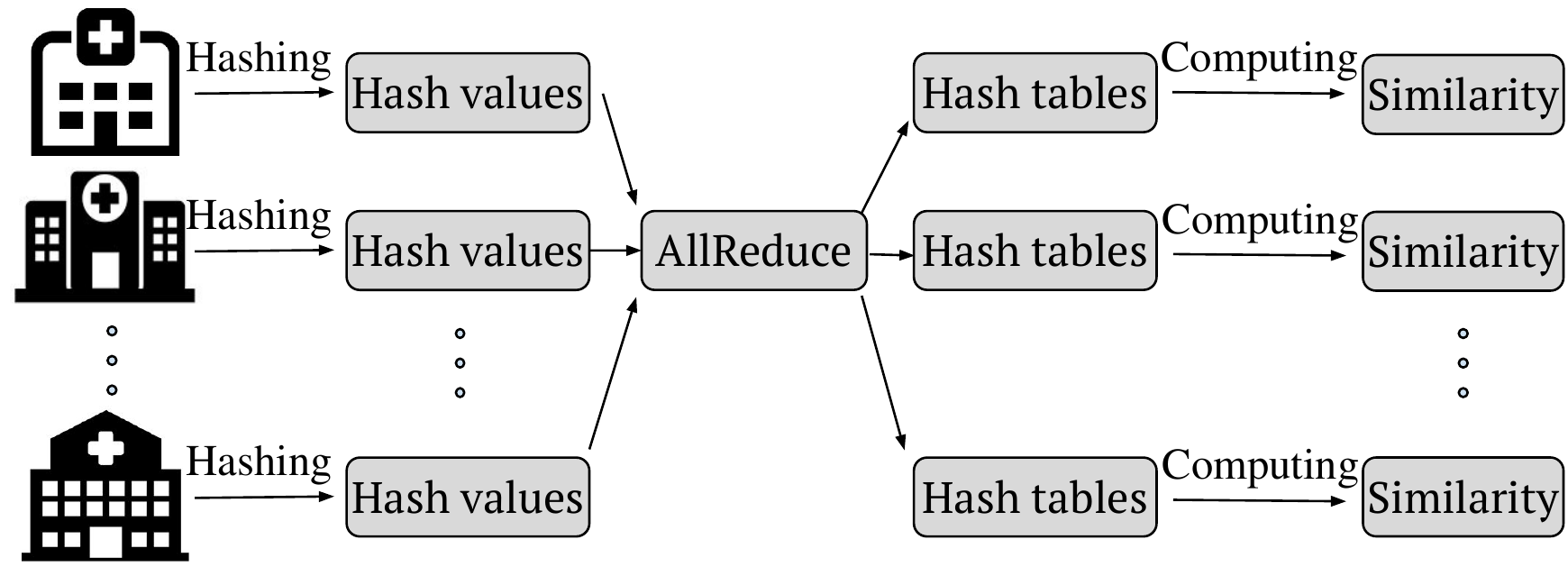}%
}
\hfil
\subfloat[The training stage]{\includegraphics[width=0.95\columnwidth]{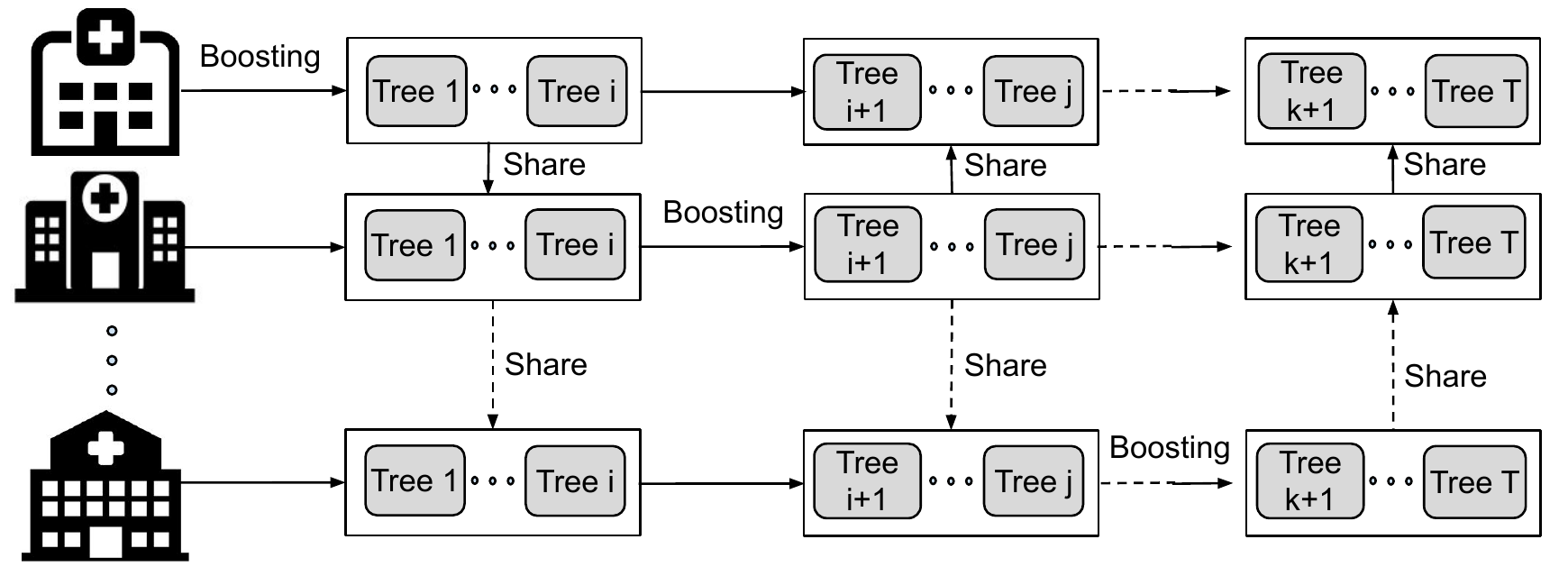}%
}
\caption{An overview of the SimFL framework}
\label{fig:protocol}
\end{figure}

\paragraph{An Overview of SimFL}
There are two main stages in SimFL: preprocessing and training. In practice, preprocessing can be done once and reuse for many runs of training. Only when the training data have updates, preprocessing has to be performed. Figure~\ref{fig:protocol} shows the structures of these two stages. 
In the preprocessing stage, each party first computes the hash values using randomly generated LSH functions. Then, by collecting the hash values from LSH, multiple global hash tables are built and broadcast to all the parties, which can be modelled as an \emph{AllReduce} communication operation~\cite{patarasuk2009bandwidth}. Finally, each party can use the global hash tables for tree building without accessing other party’s raw data.
In the training stage, all parties together train a number of trees one by one using the similarity information. Once a tree is built in a party, it will be sent to all the other parties for the update of gradients. We obtain all the decision trees as the final learned model. 

%Next, we describe the details of these two stages.

\subsection{The Preprocessing Stage}

For each instance, the aim of the preprocessing stage is to get the IDs of similar instances of the other parties. Specifically, for each party $P_m$, we want to get a matrix $\mtx{S}^m \in \mathbb{R}^{N_m \times M}$, where $\mtx{S}_{ij}^m$ is the ID of the instance in Party $P_j$ that is similar with $\mathbf{x}_i^m$. \del{In other words, for an instance $\mathbf{x}_i^m \in I_m$, the vector $\mtx{S}_{i*}^m$ contains the IDs of the instances from other parties that are similar with $\mathbf{x}_i^m$.} To obtain the similarity of any two instances in the joint data without exposing the raw data to the other parties, we adopt the widely used p-stable LSH function~\cite{datar2004locality}. According to LSH, if two instances are similar, they have a higher probability to be hashed to the same value. Thus, by applying multiple LSH functions, the bigger the number of identical hash values of two instances, the more likely they are to be similar~\cite{gionis1999similarity}.

Algorithm~\ref{alg:preprocessing} shows the process of the preprocessing stage. Given $L$ randomly generated p-stable hash functions, each party first computes the hash values of its instances. Then we build $L$ global hash tables using an AllReduce operation, which has been well studied in the previous study~\cite{patarasuk2009bandwidth}. Here the inputs to AllReduce are the instance IDs and their hash values of all parties. The reduction operation is to union the instances IDs with the same hash value. We adopt the bandwidth optimal and contention free design from the previous study~\cite{patarasuk2009bandwidth}. After broadcasting the hash tables, each party can compute the similarity information. Specifically, in party $P_m$, given an instance $\mathbf{x}_i^m$, the similar instance in the other party $P_j$ is the one with the highest count of identical hash values.\del{ If the highest count is zero, then there is no similar instance.} If there are multiple instances with the same highest count, we randomly choose one as the similar instance. In this way, the data distribution of the other parties can be learned by adopting our weighted gradient boosting strategy, which we will show next.

\begin{algorithm}
\SetNoFillComment
% \SetAlgoLined
\DontPrintSemicolon
\KwIn{LSH functions $\{\mathcal{F}_k\}_{k=1,2...L}$, Instance set $I$}
\KwOut{The similarity matrices $\mtx{S}$}
\For {each party $P_m$}{
\tcc{Conduct on party $P_m$}
    \For {each instance $\mathbf{x}_i^m \in I_m$}{
        Compute hash values $\{\mathcal{F}_k(\mathbf{x}_i^m)\}_{k=1,2...L}$;\\
    }
}
Building and broadcasting global hash tables by AllReduce;\\
\For {$m \leftarrow 1$ \KwTo $M$}{
\tcc{Conduct on party $P_m$}
    \For {each instance $\mathbf{x}_i^m \in I_m$}{
        \For {$j = 1$ \KwTo $M$}{
        \tcc{Collect similar instance IDs of $P_j$}
            \If {$j \neq m$}{
                Find the instance ID $t$ with the highest count of identical hash values;\\
                $\mtx{S}_{ij}^m \leftarrow t$;\\
            }
            \Else {
            \tcc{In the same party, the most similar instance is itself}
                $\mtx{S}_{ij}^m \leftarrow i$;\\
            }
        }
    }
}
\caption{The preprocessing stage}
\label{alg:preprocessing}
\end{algorithm}

\subsection{The Training Stage}
\label{sec:training}
In the training stage, each party trains a number of trees sequentially. When party $P_m$ is training a tree, to protect the individual records of other parties, only the local instances $I_m$ are used to learn the tree. The learned trees are shared among the parties during the training process. To exploit the similarity information between the instances from different parties, we propose a new approach to build the decision tree, which is called Weighted Gradient Boosting (WGB). The basic idea is that an instance is important and representative if it is similar to many other instances. Since the gradients can represent the importance of an instance as shown in the previous studies~\cite{chen2016xgboost,Ke2017LightGBMAH}, we propose to use weighted gradients in the training. Next, we describe WGB in detail.

\subsubsection{The Weighted Gradient Boosting Approach} 
For an instance $\mathbf{x}_q^m\in I_m$, let $\mathbf{W}_{mq}^n=\{k|\mtx{S}_{km}^n=q\}$, which contains the IDs of the instances in $I_n$ that are similar with $\mathbf{x}_q^m$. Let $g_q^m$ and $h_q^m$ denote the first order and second order gradients of loss function at $\mathbf{x}_q^m$, respectively. When $P_m$ is building a new tree at the $t$-th iteration, WGB minimizes the following objective function. 
\begin{equation}
\begin{aligned}
    &\mathcal{\tilde{L}}_w^{(t)} = \sum_{\mathbf{x}_q^m\in I_m} [\mathbf{G}_{mq}f_t(\mathbf{x}_q^m)+\frac{1}{2}\mathbf{H}_{mq}f_t^2(\mathbf{x}_q^m)]+\Omega(f_t)\\
    &\text{where } \mathbf{G}_{mq} = \sum_{n}\sum_{i\in \mathbf{W}_{mq}^n}g_i^n, \mathbf{H}_{mq} = \sum_{n}\sum_{i\in \mathbf{W}_{mq}^n}h_i^n
    % &\text{where } G_{qm} = \sum_n g_{_{\mtx{S}_{qm}^n}}, h_{w_{qm}} = \sum_n h_{\mtx{S}_{qm}^n}
\label{eq:app_loss}
\end{aligned}
\end{equation}

Compared with the objective in Eq.~\eqref{eq:xg_loss}, Eq.~\eqref{eq:app_loss} only uses the instances of $I_m$. Instead of using the gradients $g_q^m, h_q^m$ of the instance $\mathbf{x}_q^m$,  we use $\mathbf{G}_{mq}, \mathbf{H}_{mq}$ which are the sum of the gradients of the instances that are similar with $\mathbf{x}_q^m$ (including $\mathbf{x}_q^m$ itself). To help understanding, here is an example. Suppose we have two parties $P_a$ and $P_b$. When computing the similarity information for a party $P_a$, the similar instance for both $\mathbf{x}_1^a$ and $\mathbf{x}_2^a$ may be $\mathbf{x}_3^b$. Then, when building trees in $P_b$, the gradient of $\mathbf{x}_3^b$ is replaced by the aggregated gradients of $\mathbf{x}_1^a$, $\mathbf{x}_2^a$, and $\mathbf{x}_3^b$. Considering $\mathbf{G}_{mq}, \mathbf{H}_{mq}$ as \textit{weighted gradients}, we put more weights on instances that have a larger number of similar instances to utilize the similarity information.

Since the process of building a tree is similar between different parties, we only describe the process of building a tree in Party $P_m$, which is shown in Algorithm~\ref{alg:boost_tree}. At first, the parties update the gradients of the local instances. Then, for each instance of $P_m$, the other parties compute and send the aggregated gradients of the similar instances. Instead of sending each gradient directly, such aggregation on the local party can reduce the communication cost and protect the individual gradients. After all the aggregated gradients are computed and send to $P_m$, the weighted gradients can be easily computed by summing the aggregated gradients. Then, we can build a tree based on the weighted gradients.

\begin{algorithm}[t]
\SetNoFillComment
% \SetAlgoLined
\DontPrintSemicolon
\KwIn{Instance set $I$}
\KwOut{A new decision tree}
% \tcc{Enumerate parties $P_1$ to $P_M$ except $P_m$}
\For{$i = 1$ \KwTo $M$, $i \neq m$}{
\tcc{Conduct on party $P_i$}
    $\mathbf{G}_{m*}^i \leftarrow \mathbf{0}$, $\mathbf{H}_{m*}^i \leftarrow \mathbf{0}$;\\
    Update the gradients of instances in $I_i$;\\
    \For{each instance $\mathbf{x}_q^i \in I_i$}{
        Get the similar instance ID $s=\mathbf{S}_{qm}^i$;\\
        $\mathbf{G}_{ms}^i \leftarrow \mathbf{G}_{ms}^i +  g_q^i$, $\mathbf{H}_{ms}^i \leftarrow \mathbf{G}_{ms}^i +  h_q^i $;\\
        
    }
    Send $\mathbf{G}_{m*}^i$, $\mathbf{H}_{m*}^i$ to $P_m$;
}
\tcc{Conduct on party $P_m$}
Update the gradients of instances in $I_m$;\\
\For{each instance $\mathbf{x}_q^m \in I_m$}{
    $\mathbf{G}_{mq} \leftarrow 0$, $\mathbf{H}_{mq} \leftarrow 0$;\\
    \For{$i \leftarrow 1$ \KwTo $M$}{
        \If{$i == m$}{
            $\mathbf{G}_{mq} \leftarrow \mathbf{G}_{mq} + g_q^m$;\\
            $\mathbf{H}_{mq} \leftarrow \mathbf{H}_{mq} + h_q^m$;
        }
        \Else{
            $\mathbf{G}_{mq} \leftarrow \mathbf{G}_{mq} + \mathbf{G}_{mq}^i$;\\
            $\mathbf{H}_{mq} \leftarrow \mathbf{H}_{mq} + \mathbf{H}_{mq}^i$;
        }
    }
}
Build a tree with instances $I_m$ and weighted gradients $\mathbf{G}_{m*}$, $\mathbf{H}_{m*}$;\\
Send the tree to the other parties;
\caption{The process of learning a tree}
\label{alg:boost_tree}
\end{algorithm}

\section{Theoretical Analysis}

% In this section, we analyze the privacy level, utility and efficiency of our proposed approach.

\subsection{Privacy Level Analysis}

\begin{theorem}
\label{theorem:secure}

The SimFL protocol satisfies the privacy model definition if $L < d$, where $L$ is the number of hash functions and $d$ is the number of dimensions of training data.
\end{theorem}

In short, there are infinite number of solutions if the number of unknowns (i.e., $d$) is bigger than the number of equations (i.e., $L$)~\cite{ladyzhenskaia1968linear}. The detailed proof is available in Appendix A.
% of the supplementary material. 

Theorem~\ref{theorem:secure} indicates that, when $L<d$, SimFL ensures that, for any its output, there exists an infinite number of possible inputs resulting the same output. Therefore, a dishonest party cannot determine the actual raw data from other parties. Potentially, there may be background knowledge attack against SimFL. An optional method to resolve this problem is to decrease $L$ to get a fewer number of equations, which results in a larger number of inputs for the same output. For example, if we know the background knowledge of a feature and now we only have $(d-1)$ unknowns, we can set $L$ to $(d-2)$ or even smaller to ensure that the raw data  still cannot be extracted. 

\subsection{The Error of Weighted Gradient Boosting}
\label{sec:error_formula}
Here we analyze the approximation error as well as the generalization error of WGB.
\begin{theorem}
\label{theorem:upperbound}
For simplicity, we assume that the feature values of each dimension are i.i.d. uniform random variables and the split value is randomly chosen from the feature values. Suppose $P_m$ is learning a new tree. We denote the approximation error of WGB as $\varepsilon^t = |\mathcal{\tilde{L}}_w^{(t)} - \mathcal{\tilde{L}}^{(t)}|$. Let $d_t = \max_{r,j}\norm{\mathbf{x}_{\mtx{S}_{rm}^j}^m-\mathbf{x}_r^j}_1$, $d_{m} = \max_{i,j}\norm{\mathbf{x}_i- \mathbf{x}_j}_1$, $g'=\max_i |g_i|$, $h'=\max_i |h_i|$, and $f_t'=\max_i |f_t(\mathbf{x}_i)|$. Then, with probability at least $1-\delta$, we have 
\begin{equation}
\begin{aligned}
\label{eq:upperbound}
    \varepsilon^t \leq &\Big([1-(1-\frac{d_t}{d_m})^D](N-N_m)+\sqrt{\frac{(N-N_m)\ln{\frac{1}{\delta}}}{2}}\Big) \\
    & \cdot (2g' f_t'+\frac{1}{2}h' f_t'^2)  
\end{aligned}
\end{equation}
where $D$ is the depth of the tree, $N$ is the number of instances in $I$, and $N_m$ is the number of instances in $I_m$.
\end{theorem}

The detailed proof of Theorem~\ref{theorem:upperbound} is available in Appendix B. According to Theorem~\ref{theorem:upperbound}, the upper bound of the approximation error is $\mathcal{O}(N-N_m)$ with respect to the number of instances. The approximation error of WGB may increase as the number of instances of the other parties increases. However, let us consider the generalization error of WGB, which can be formulated as $\varepsilon_{gen}^{WGB}(t) = |\mathcal{\tilde{L}}_w^{(t)}-\mathcal{\tilde{L}}_*^{(t)}| \leq |\mathcal{\tilde{L}}_w^{(t)} - \mathcal{\tilde{L}}^{(t)}| + |\mathcal{\tilde{L}}^{(t)}-\mathcal{\tilde{L}}_*^{(t)}|\triangleq \varepsilon^t + \varepsilon_{gen}(t)$. The generalization error of vanilla GBDTs (i.e., $\varepsilon_{gen}(t)$) tends to decrease as the number of training instances $N$ increases~\cite{shalev2014understanding}. Thus, WGB may still have a low generalization error as $N$ increases. 
% As we will show in the experiments, SimFL has a very good performance for both small and large datasets.

\subsection{Computation and Communication Efficiency}
Here we analyze the computation and communication overhead of SimFL. Suppose we have $T$ tress, $M$ parties, $N$ total training instances, and $L$ hash functions.

\paragraph{Computation Overhead} (1) In the preprocessing stage, we first need to compute the hash values, which costs $O(Nd)$. Then, we need to union the instance IDs with the same hash value and compute the similarity information. The union operation can be done in $\mathcal{O}(NL)$ since we have to traverse $NL$ hash values in total. When computing the similarity information for an instance $\mathbf{x}_i^m$, a straightforward method is to union the instance IDs of $L$ buckets with hash values $\{\mathcal{F}_k(\mathbf{x}_i^m)\}_{k=1,2...L}$ and conduct linear search to find the instance ID with the maximum frequency. On average, each hash bucket has a constant number of instances. Thus, for each instance, the linear search can be done in $\mc{O}(L)$. Thus, the total computation overhead in the preprocessing stage is $\mc{O}(NL+Nd)$ on average.

(2) In the training stage, the process of building a tree is the same as the vanilla GBDTs except computing the weighted gradients. The calculation of the weighted gradients is done by simple sum operations, which is $\mathcal{O}(N)$. Then, the computation overhead is $\mathcal{O}(NT)$ in the training.

\paragraph{Communication Overhead} Suppose each real number uses 4 bytes to store. (1) In the preprocessing stage, according to the previous study~\cite{patarasuk2009bandwidth}, since we have $NL$ hash values and the corresponding instance IDs to share, the total communication cost in the AllReduce operation is $8MNL$ bytes. (2) In the training stage, each party has to send the aggregated gradients. The size of the gradients is no more than $8N$ (including $g$ and $h$). After building a tree, $P_m$ has to send the tree to the other parties. Suppose the depth of each tree is $D$. Our implementation uses 8 bytes to store the split value and the feature ID of each node. Therefore, the size of a tree is $8(2^D-1)$. Since each tree has to be sent to $(M-1)$ parties, the communication cost for one tree is $8(2^D-1)(M-1)$. The total communication overhead for a tree is $8[N +(2^D-1)(M-1)]$. Since there are $T$ trees, the communication overhead in the training stage is $8T[N +(2^D-1)(M-1)]$.

\section{Experiments}
We present the effectiveness and efficiency of SimFL. To understand the model accuracy of SimFL, we compare SimFL with two approaches: 1) {\bf SOLO}: Each party only trains vanilla GBDTs with its local data. This comparison shows the incentives of using SimFL. 2) {\bf ALL-IN}: A party trains vanilla GBDTs with the joint data from all parties without the concern of privacy. This comparison demonstrates the potential accuracy loss of achieving the privacy model. We also compare SimFL with the distributed boosting framework proposed by~\citet{zhao2018inprivate} (referred as {\bf TFL} (Tree-based Federated Learning)). Here we only adopt their framework and do not add any noise in the training (their system also adds noises to training for differential privacy). 

We conducted the experiments on a machine running Linux with two Xeon E5-2640v4 10 core CPUs, 256GB main memory and a Tesla P100 GPU of 12GB memory. To take advantage of GPU, we use ThunderGBM~\cite{ThunderGBM} in our study. We use six public datasets from the LIBSVM website\footnote{\url{https://www.csie.ntu.edu.tw/~cjlin/libsvm/index.html}}, as listed in Table~\ref{tbl:dataset}. We use 75\% of the datasets for training and the remainder for testing. The maximum depth of the trees is set to 8. For the LSH functions, we choose $r = 4.0$ and $L=\min\{40, d-1\}$, where $d$ is the dimension of the dataset. The total number of trees is set to 500 in all approaches. Due to the randomness of the LSH functions, the results of SimFL may differ in the different runnings. We run SimFL for 10 times in each experiment and report the average, minimal and maximum errors.

\begin{table}
\centering
\caption{datasets used in the experiments}
\label{tbl:dataset}
\begin{tabular}{|c|c|c|c|}
\hline
dataset  & cardinality & dimension &data size\\ \hline
a9a    & 32,561      & 123   &16MB    \\ \hline
cod-rna      & 59,535      & 9  &2.1MB     \\ \hline
real-sim & 72,309 &20,958 &6.1GB\\ \hline
ijcnn1 & 49,990 & 22 &4.4MB\\ \hline
SUSY & 1,000,000 & 18 &72MB\\ \hline
HIGGS  & 1,000,000 &28 &112MB\\ \hline
\end{tabular}
\end{table}

In reality, the distribution of the data in different parties may vary. For example, due to the ozone hole, the countries in the Southern Hemisphere may have more skin cancer patients than the Northern Hemisphere. Thus, like a previous work~\cite{pmlr-v97-yurochkin19a}, we consider two ways to divide the training dataset to simulate different data distributions in the federated setting: \emph{unbalanced partition} and \emph{balanced partition}. In the unbalanced partition, we divide the datasets with two classes (i.e., 0 and 1) to subsets where each subset has a relatively large proportion of the instances in one class. Given a ratio $\theta \in (0,1)$, we randomly sample $\theta$ proportion of the instances that are with the class 0 and $(1-\theta)$ proportion of the instances that are with the class 1 to form a subset, and the remained for the other subset. Then, we can split these two subsets to more parts equally and randomly to simulate more parties. In the balanced partition, we simply split the datasets equally and randomly assign the instances to different parties.

\subsection{Test Errors}
\label{sec:error}

We first divide the datasets into two parts using the unbalanced partition with the ratio $\theta=80\%$, and assign them to two parties $A$ and $B$ . The test errors are shown in Table~\ref{tbl:err2unb}. We have the following observations. First, the test errors of SimFL are always lower than SOLO on data parts A and B (denoted as SOLO$_A$ and SOLO$_B$ respectively). The accuracy can be improved by about 4\% on average by performing SimFL. Second, the test error of SimFL is close to ALL-IN.  Third, Compared with TFL, SimFL has much lower test errors. The test errors of TFL are always bigger than SOLO, which discourages the adoptions of TFL in practice. In other words, TFL's approach on aggregating decision trees from multiple parties cannot improve the prediction of the local data of individual parties.

\begin{table}
\centering
\caption{The test errors of different approaches ($\theta =80\%$)}
\label{tbl:err2unb}
\resizebox{.95\columnwidth}{!}{%
\begin{tabular}{|c|c|c|c|c|c|c|c|}
\hline
\multirow{2}{*}{datasets} & \multicolumn{3}{c|}{SimFL}  & \multirow{2}{*}{TFL} & \multirow{2}{*}{SOLO$_A$} & \multirow{2}{*}{SOLO$_B$} & \multirow{2}{*}{ALL-IN} \\ \cline{2-4}
                           & avg     & min     & max     &                      &                      &                      &                     \\ \hline
a9a                        & 17.0\% & 16.7\% & 17.2\% & 23.1\%              & 19.1\%              & 22.0\%              & 15.1\%             \\ \hline
cod-rna                    & 6.5\%  & 6.3\%  & 6.7\%  & 7.5\%              & 9.6\%              & 8.2\%              & 6.13\%              \\ \hline
real-sim                  &8.3\% &8.2\% &8.4\% &10.1\% &11.8\% &14.4\% &6.5\%        \\ \hline
ijcnn1                   & 4.5\% &4.5\% & 4.6\% &4.7\% &5.9\% &4.9\% &3.7\%    \\ \hline
SUSY                       &  23.3\%    & 23.1 \%              & 23.4\%     & 32.6\%     & 25.4\%   &32.8 \%                      & 21.38\%             \\ \hline
HIGGS                      & 30.9\% & 30.8\% & 31.0\% & 36.1\%              & 37.1\%              & 35.6\%              & 29.4\%             \\ \hline
\end{tabular}%
}
\end{table}

\begin{table}
\centering
\caption{The test errors of different approaches (balanced partition)}
\label{tbl:err2b}
\resizebox{.95\columnwidth}{!}{%
\begin{tabular}{|c|c|c|c|c|c|c|c|}
\hline
\multirow{2}{*}{datasets}   & \multicolumn{3}{c|}{SimFL}       & \multirow{2}{*}{TFL} & \multirow{2}{*}{SOLO$_A$} & \multirow{2}{*}{SOLO$_B$} & \multirow{2}{*}{ALL-IN} \\ \cline{2-4}
                                                        & avg      & min       & max       &                      &                      &                      &                     \\ \hline
a9a                       & 15.1\%  & 14.9\%  & 15.2\%  & 15.6\%              & 15.5\%              & 15.9\%              &15.1\%            \\ \hline
cod-rna                                                 & 6.0\%  & 5.9\% & 6.1\% & 6.3\%               & 6.4\%               & 6.7\%               & 6.1\%           \\ \hline
real-sim          &7.1\% &6.9\% &7.2\% &7.4\% &8.4\% &7.9\% &6.5\%              \\ \hline
ijcnn1             &3.6\% &3.5\% &3.7\% &3.8\% &3.8\% &4.2\% &3.7\%     \\ \hline
SUSY                                                    & 19.6\% &  19.4\%        &  19.9\%          &20.2\%             & 20.1\%             & 20.1\%             & 20.0\%            \\ \hline
HIGGS                      &   29.3\%        &  29.2\%&    29.5\%              &    31.4\%      & 30.2\%          & 30.5\%          &  29.3\%                                  \\ \hline
\end{tabular}%
}
\end{table}

\begin{figure}
% \captionsetup[subfloat]{farskip=2pt,captionskip=1pt}
\centering
\subfloat[a9a]{\includegraphics[width=.475\columnwidth]{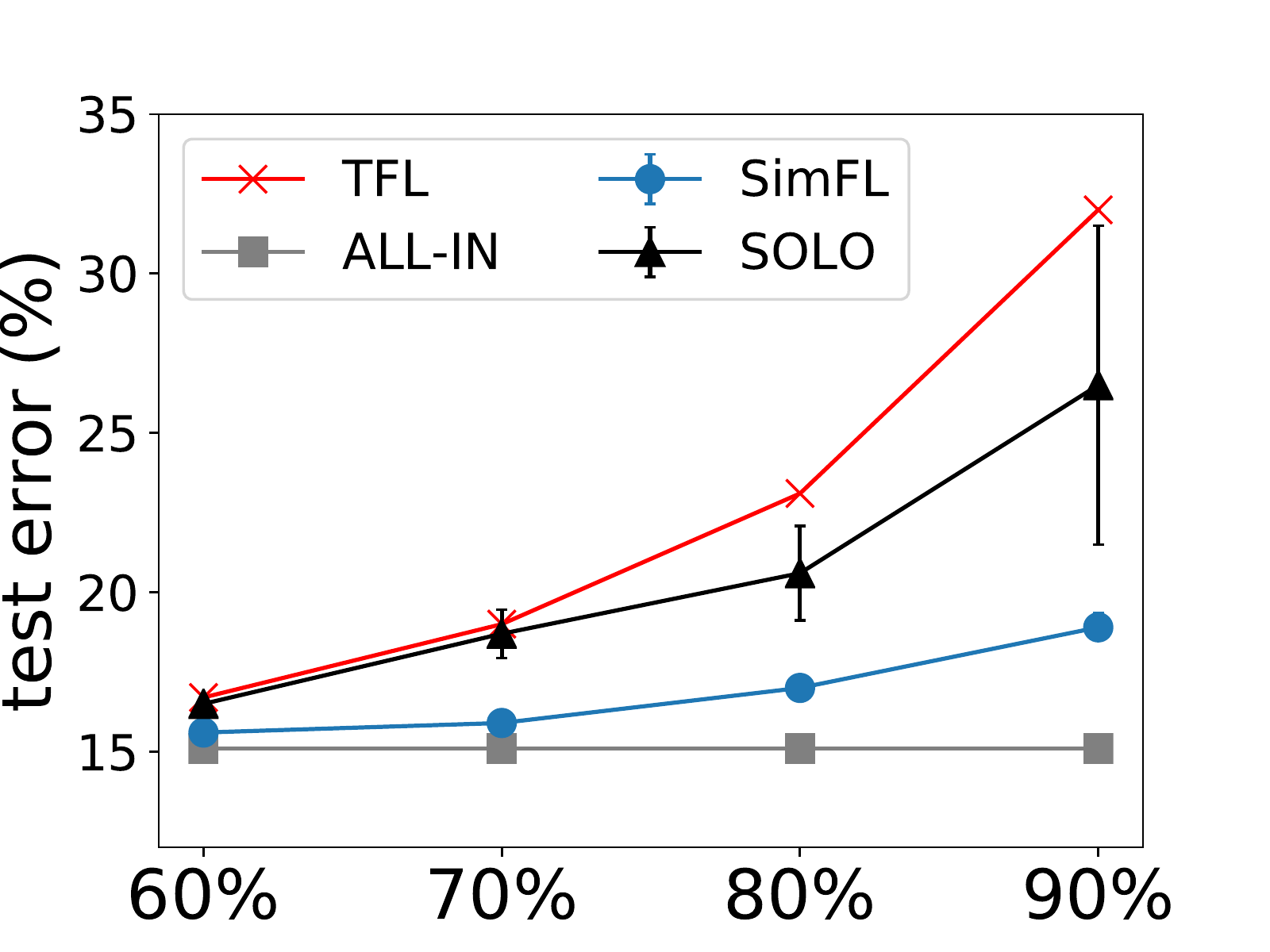}%
}
% \hfil
\subfloat[cod-rna]{\includegraphics[width=.475\columnwidth]{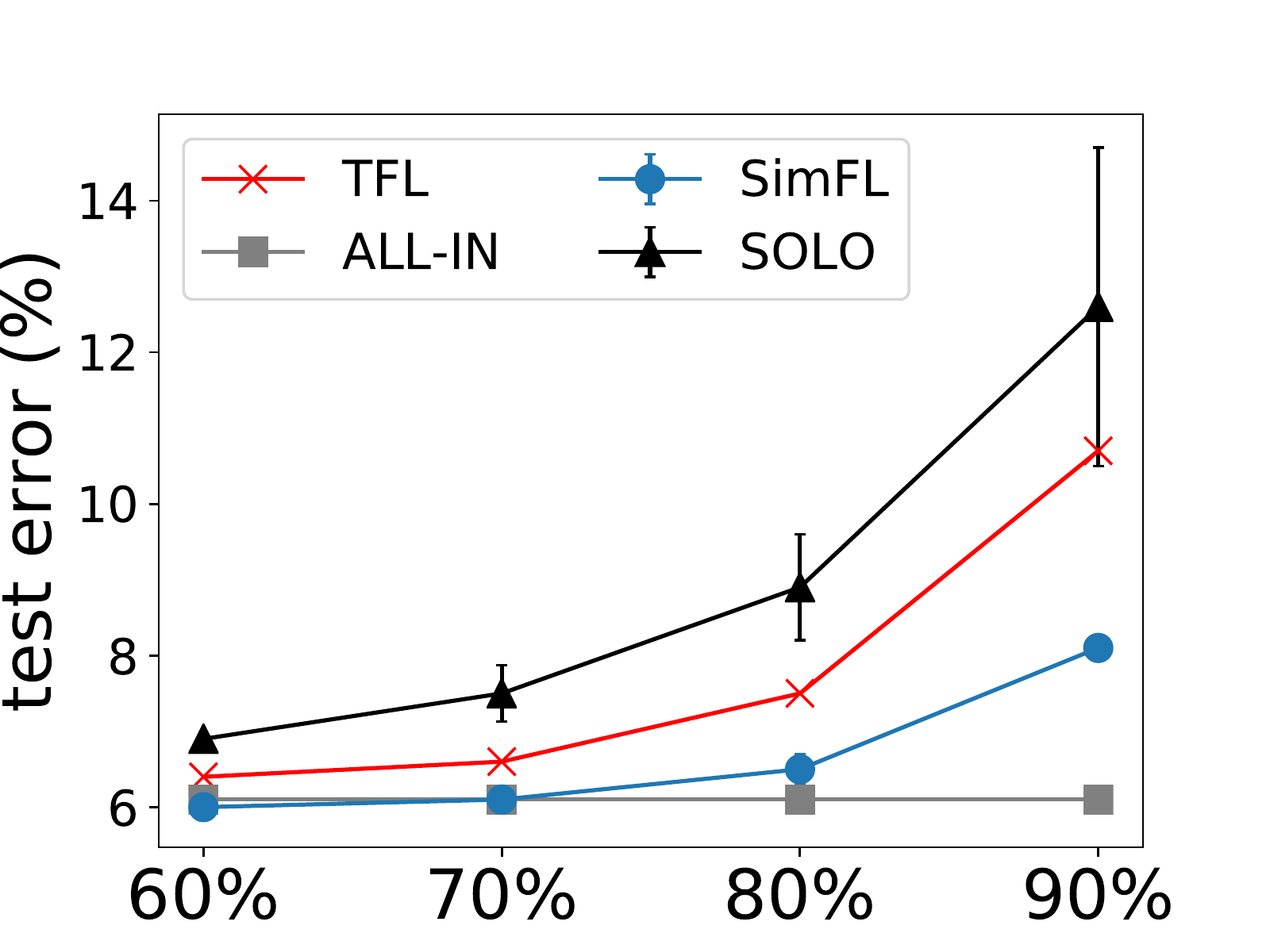}%
}
\hfil
\subfloat[real-sim]{\includegraphics[width=.475\columnwidth]{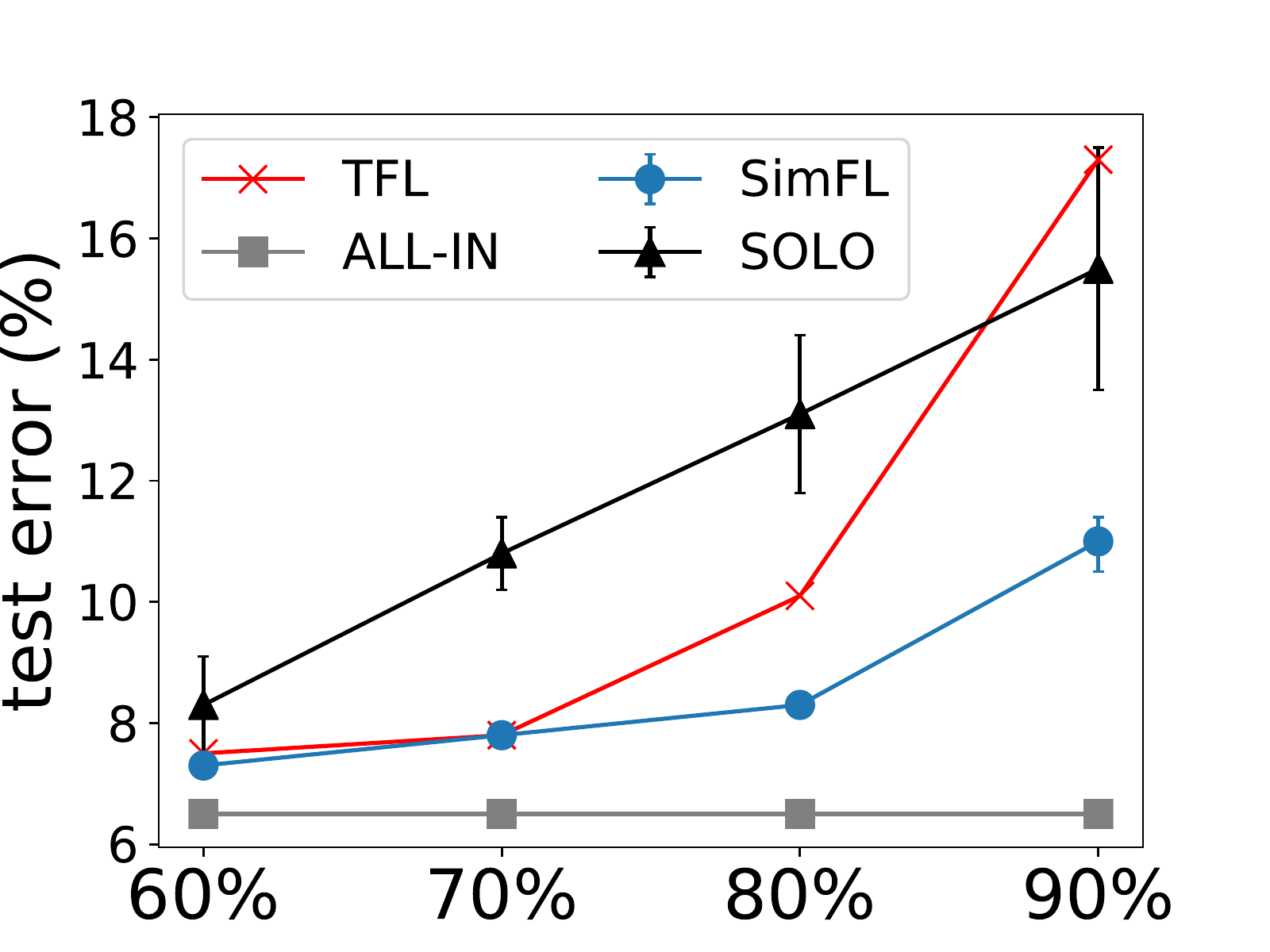}%
}
% \hfil
\subfloat[ijcnn1]{\includegraphics[width=.475\columnwidth]{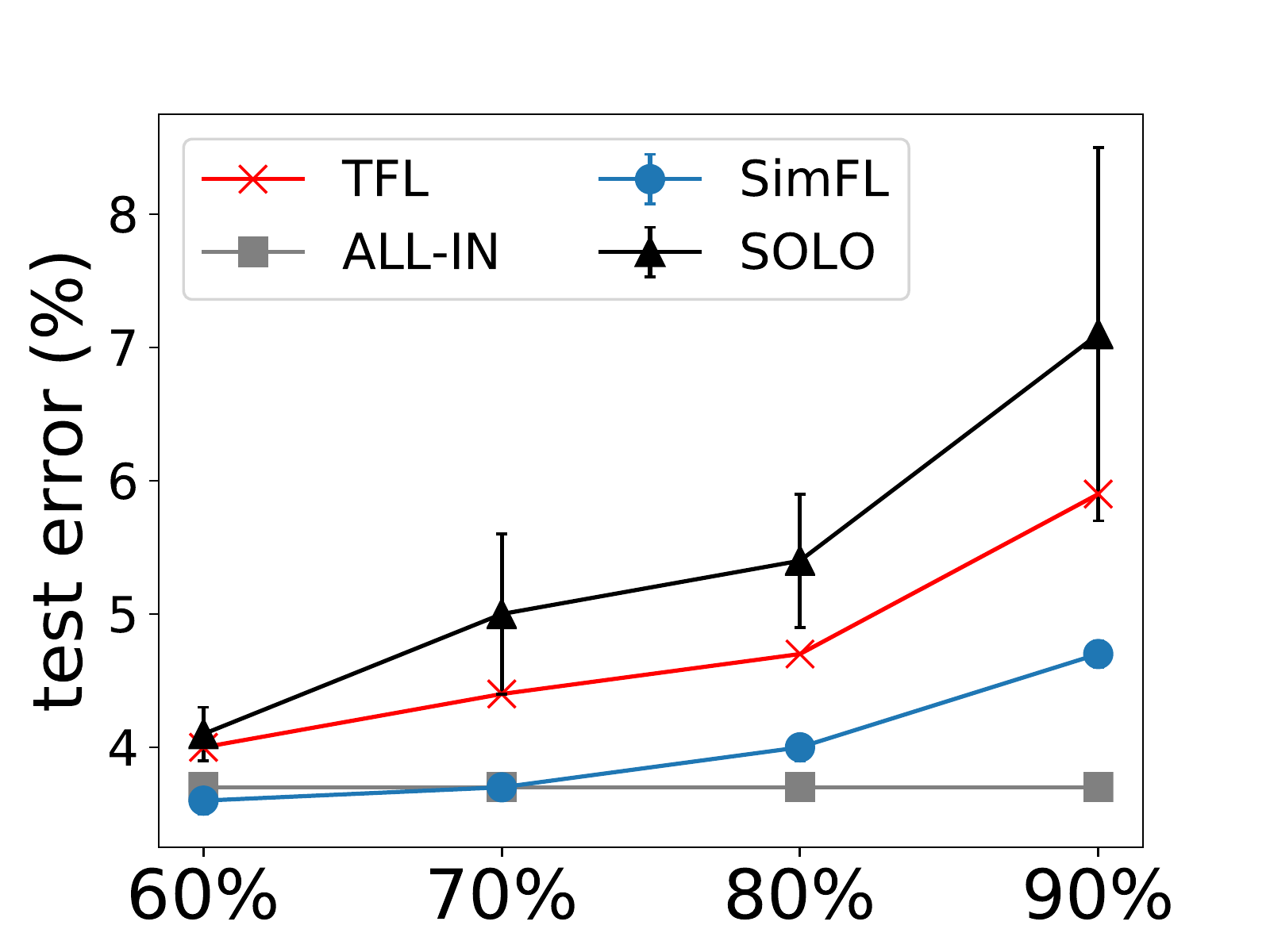}%
}
\hfil
\subfloat[SUSY]{\includegraphics[width=.475\columnwidth]{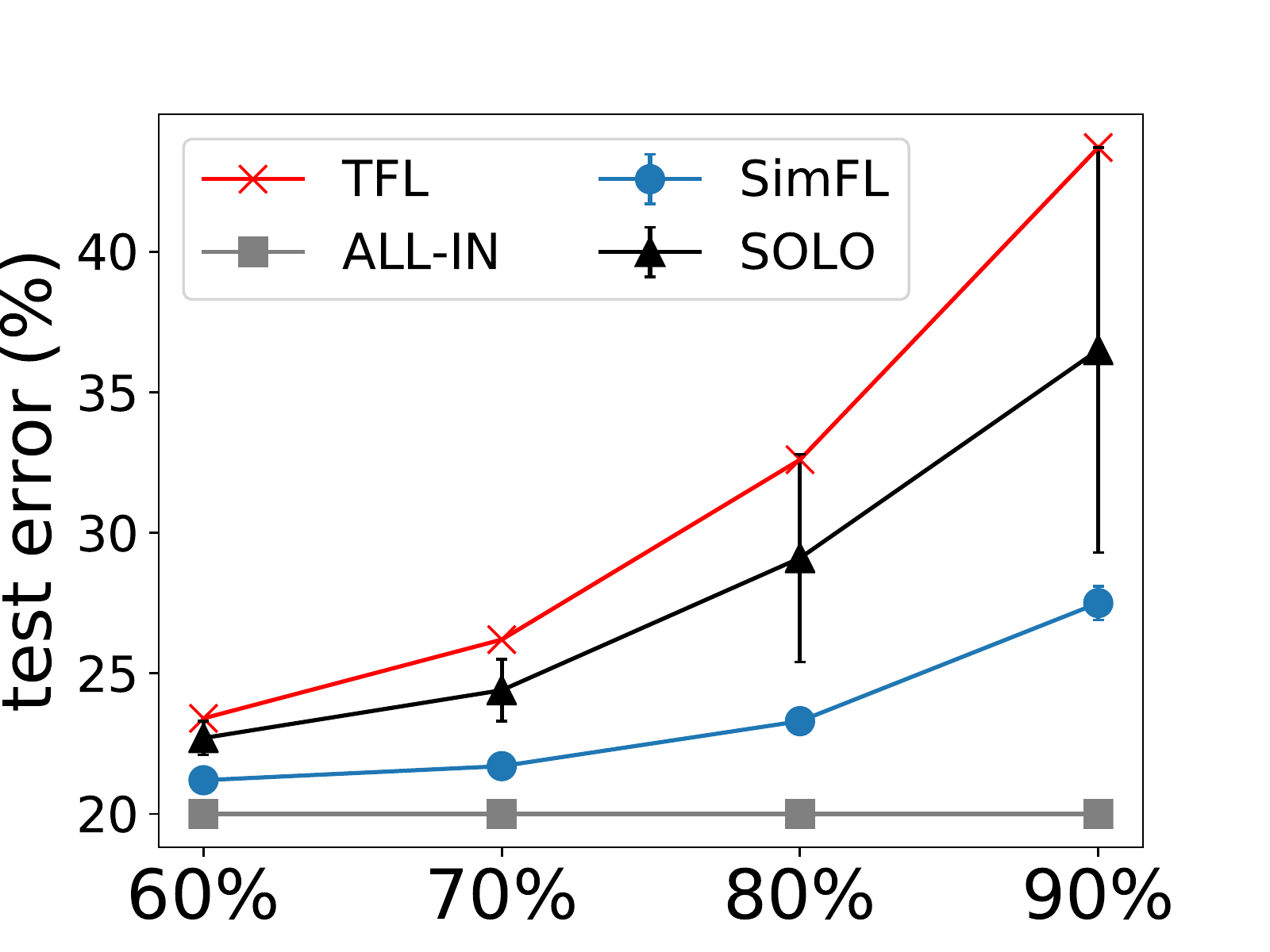}%
}
\subfloat[HIGGS]{\includegraphics[width=.475\columnwidth]{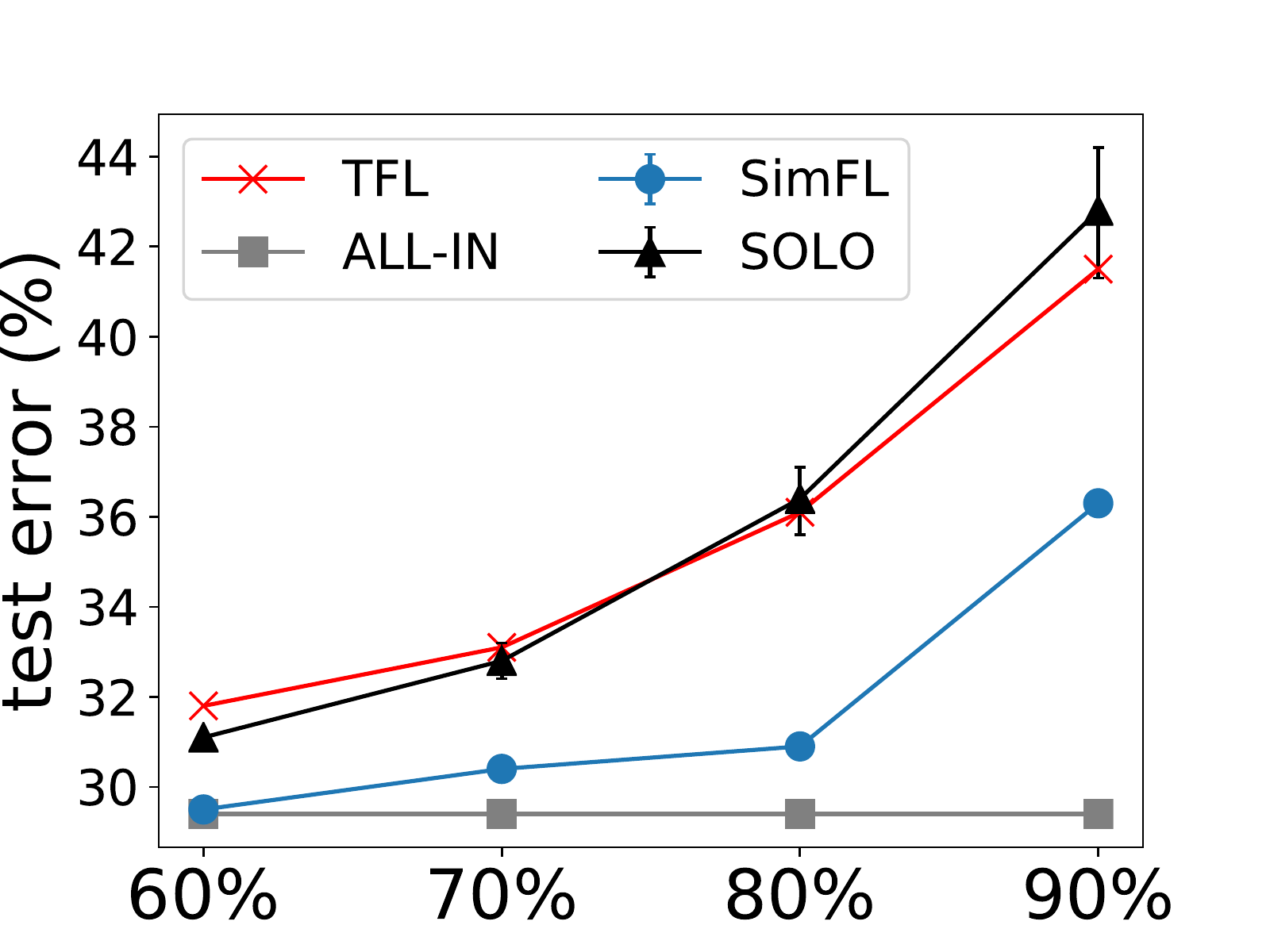}%
}
\caption{The test error with different ratio $\theta$}
\label{fig:errorratio}
\end{figure}

To figure out the impact of the ratio $\theta$, we conduct experiments with different ratios that range from $60\%$ to $90\%$. The results are shown in Figure~\ref{fig:errorratio}. Compared with TFL and SOLO, SimFL works quite well especially when $\theta$ is high. The similarity information can effectively help to discover the distribution of the joint data among multiple parties. The variation of SimFL is low in multiple runs. 

Table~\ref{tbl:err2b} shows the errors with the setting of the balanced partition. It seems that for the six datasets, the local data in each party is good enough to train the model. The test errors of SOLO and ALL-IN are very close to each other. However, our SimFL still performs better than SOLO and TFL, and sometimes even has a lower test error than ALL-IN. Even in a balanced partition, SimFL still has a good performance.

\begin{figure}
% \captionsetup[subfloat]{farskip=2pt,captionskip=1pt}
\centering
\subfloat[a9a]{\includegraphics[width=.475\columnwidth]{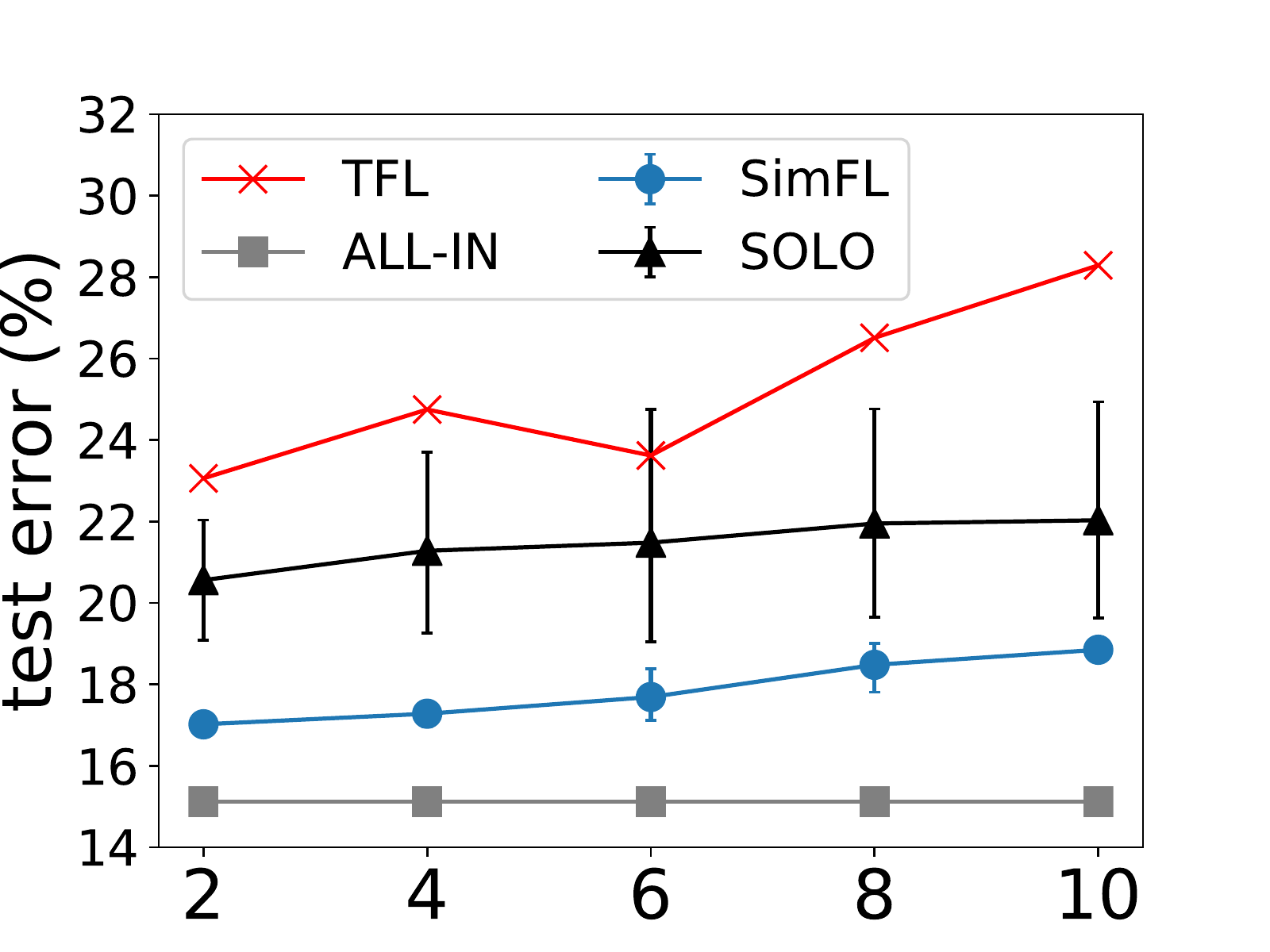}%
}
\subfloat[cod-rna]{\includegraphics[width=.475\columnwidth]{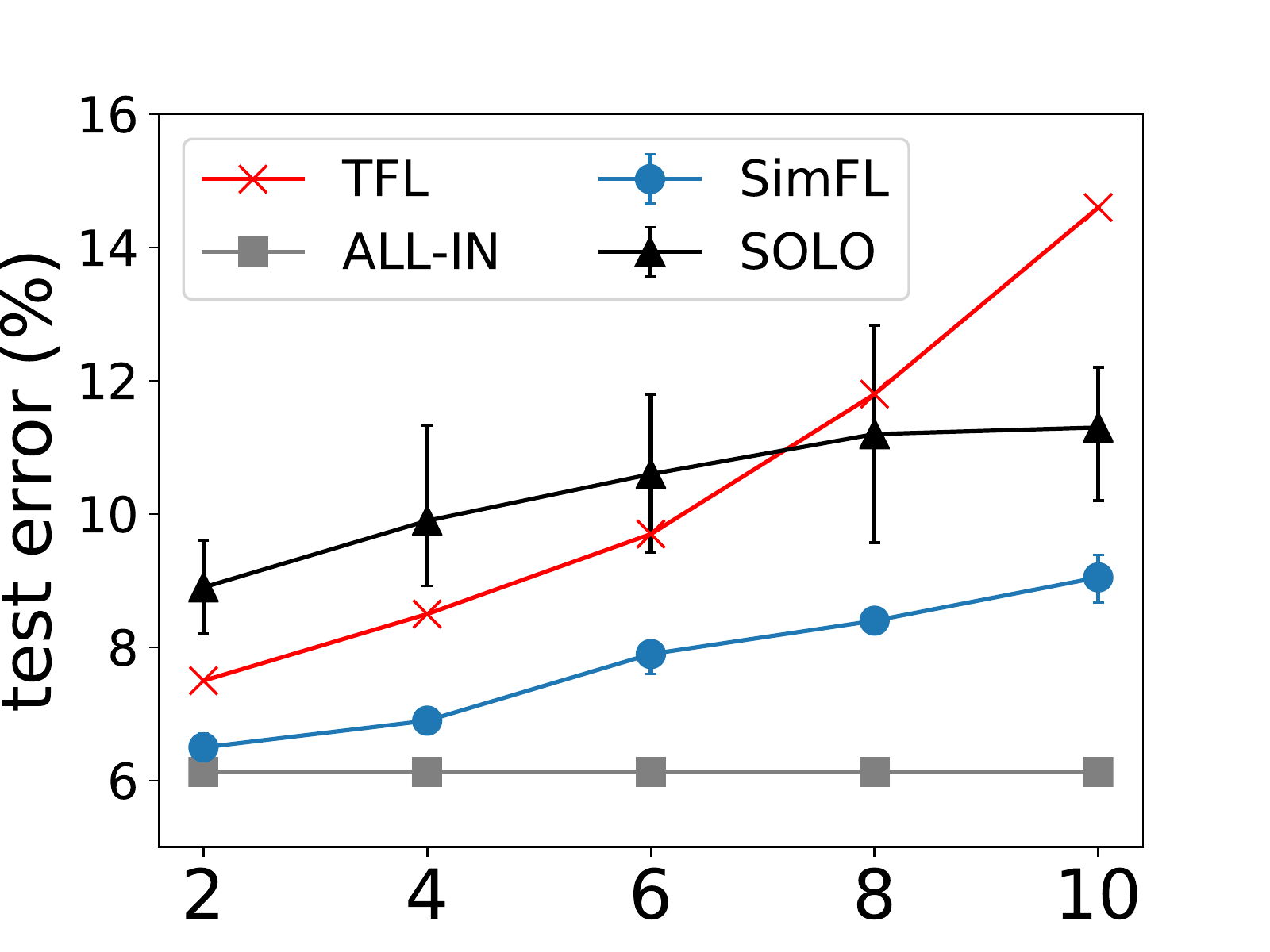}%
}
\hfil
\subfloat[real-sim]{\includegraphics[width=.475\columnwidth]{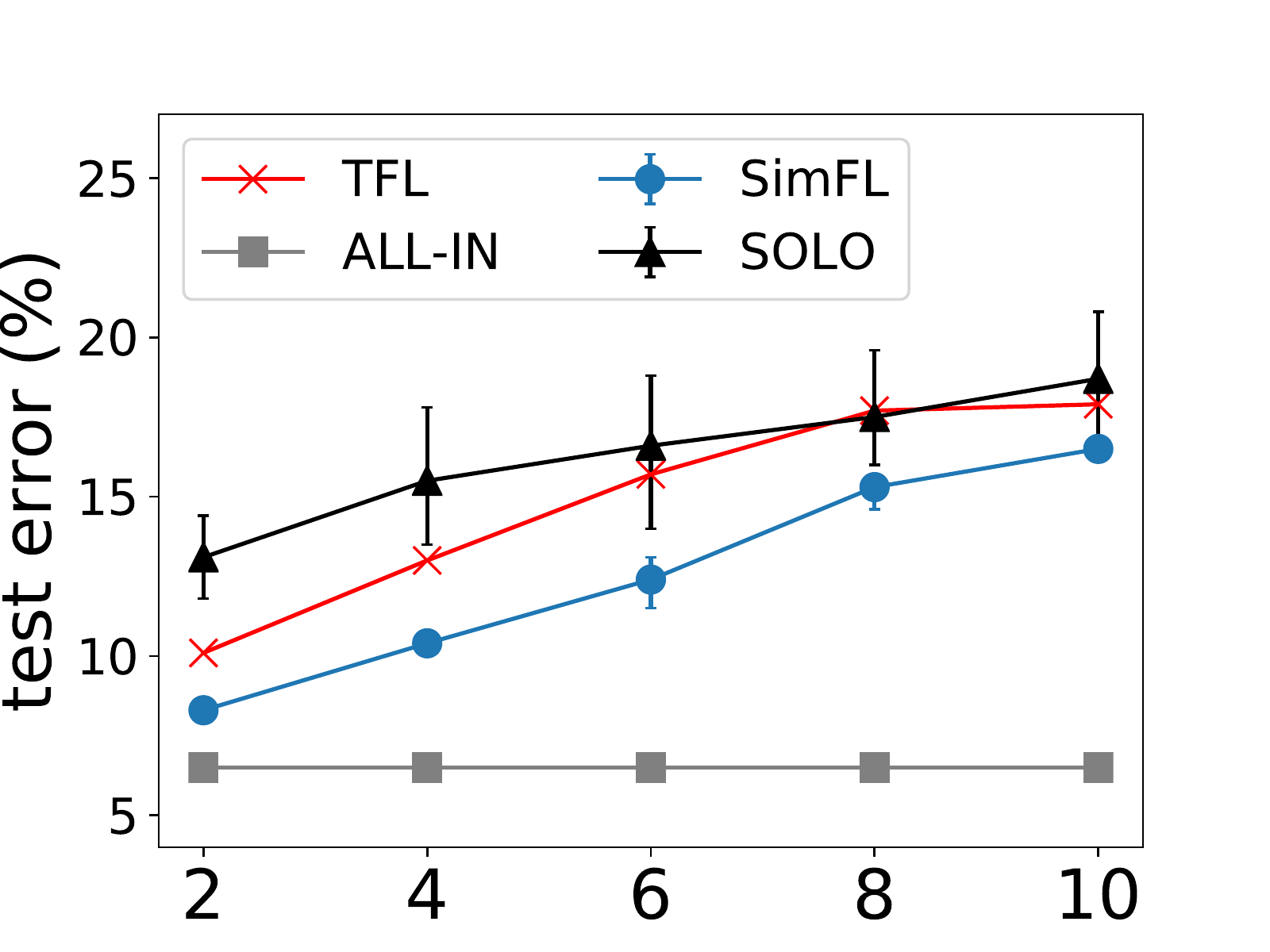}%
}
\subfloat[ijcnn1]{\includegraphics[width=.475\columnwidth]{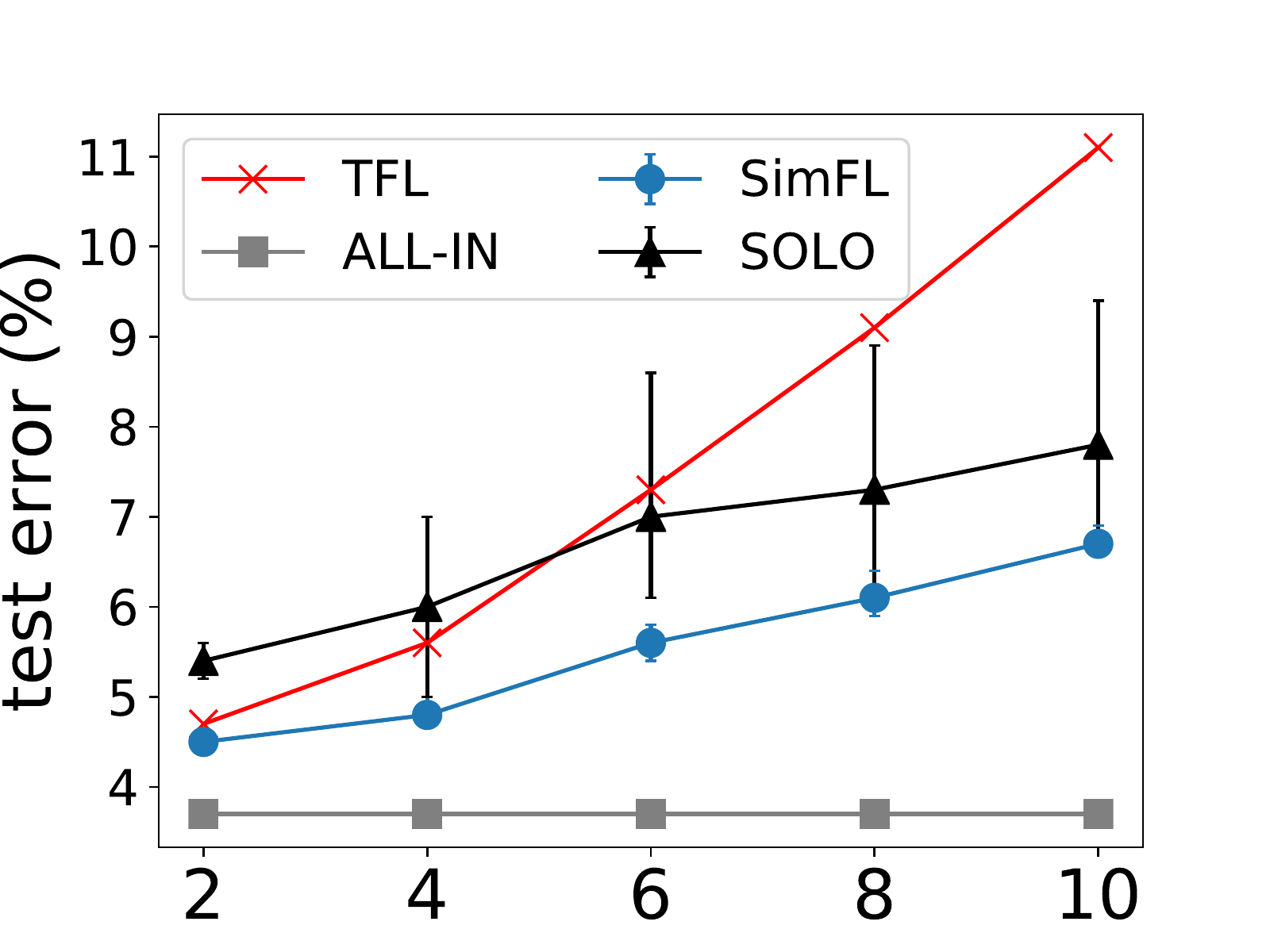}%
}
\hfil
\subfloat[SUSY]{\includegraphics[width=.475\columnwidth]{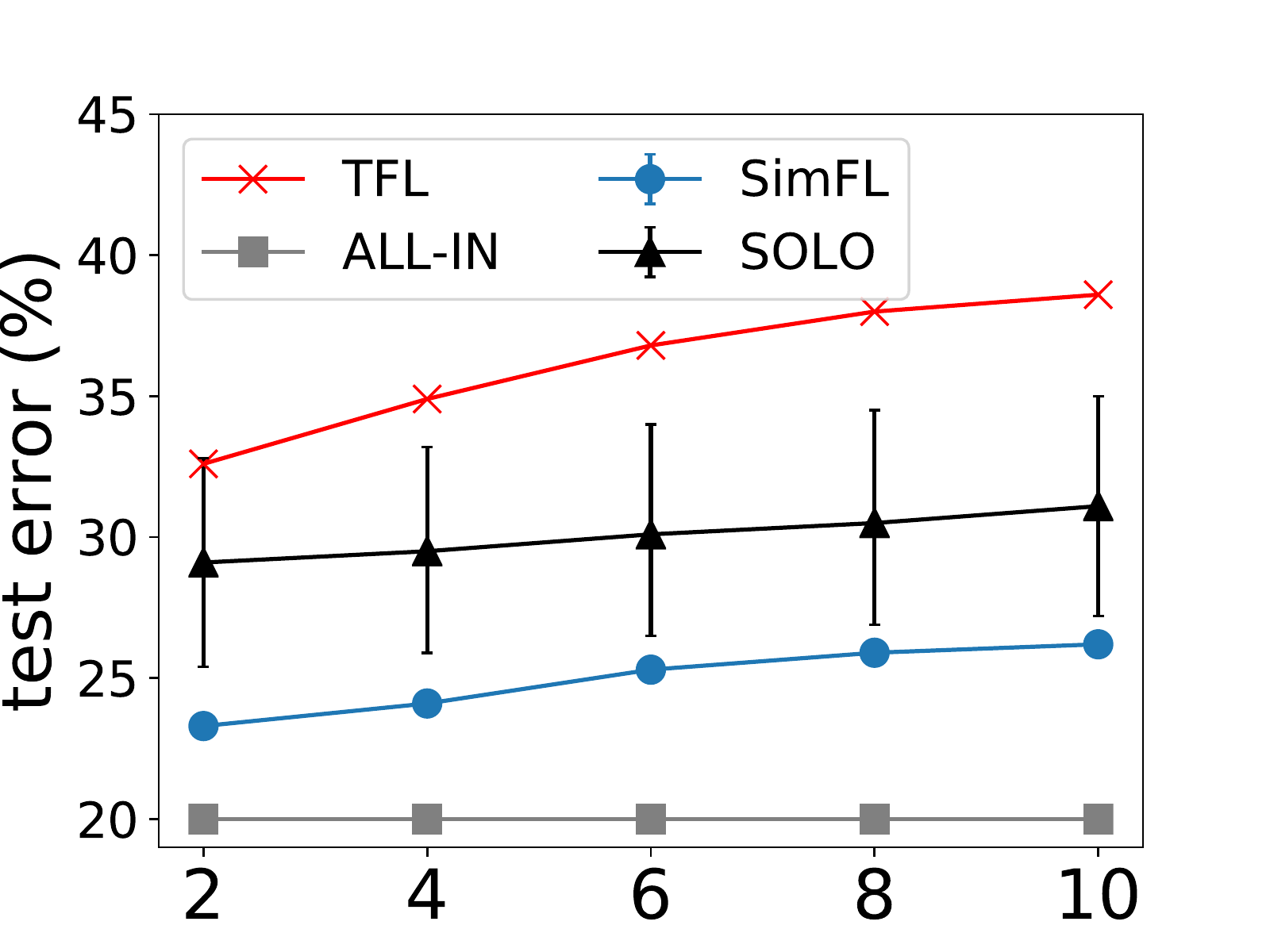}%
}
\subfloat[HIGGS]{\includegraphics[width=.475\columnwidth]{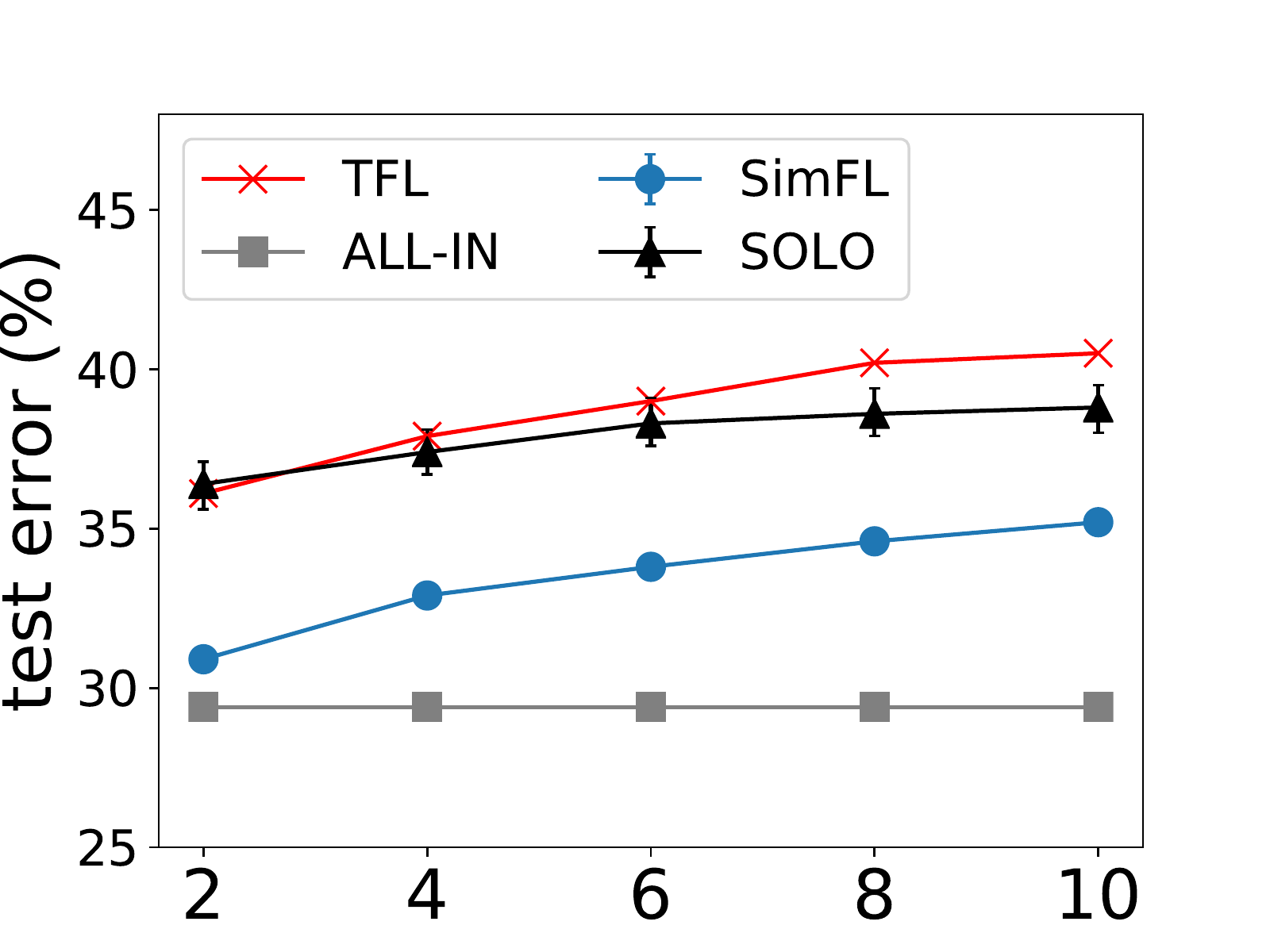}%
}
\caption{The impact of number of parties ($\theta=80\%$)}
\label{fig:errorubnp}
\end{figure}

\begin{figure}
% \captionsetup[subfloat]{farskip=2pt,captionskip=1pt}
\centering
\subfloat[a9a]{\includegraphics[width=.475\columnwidth]{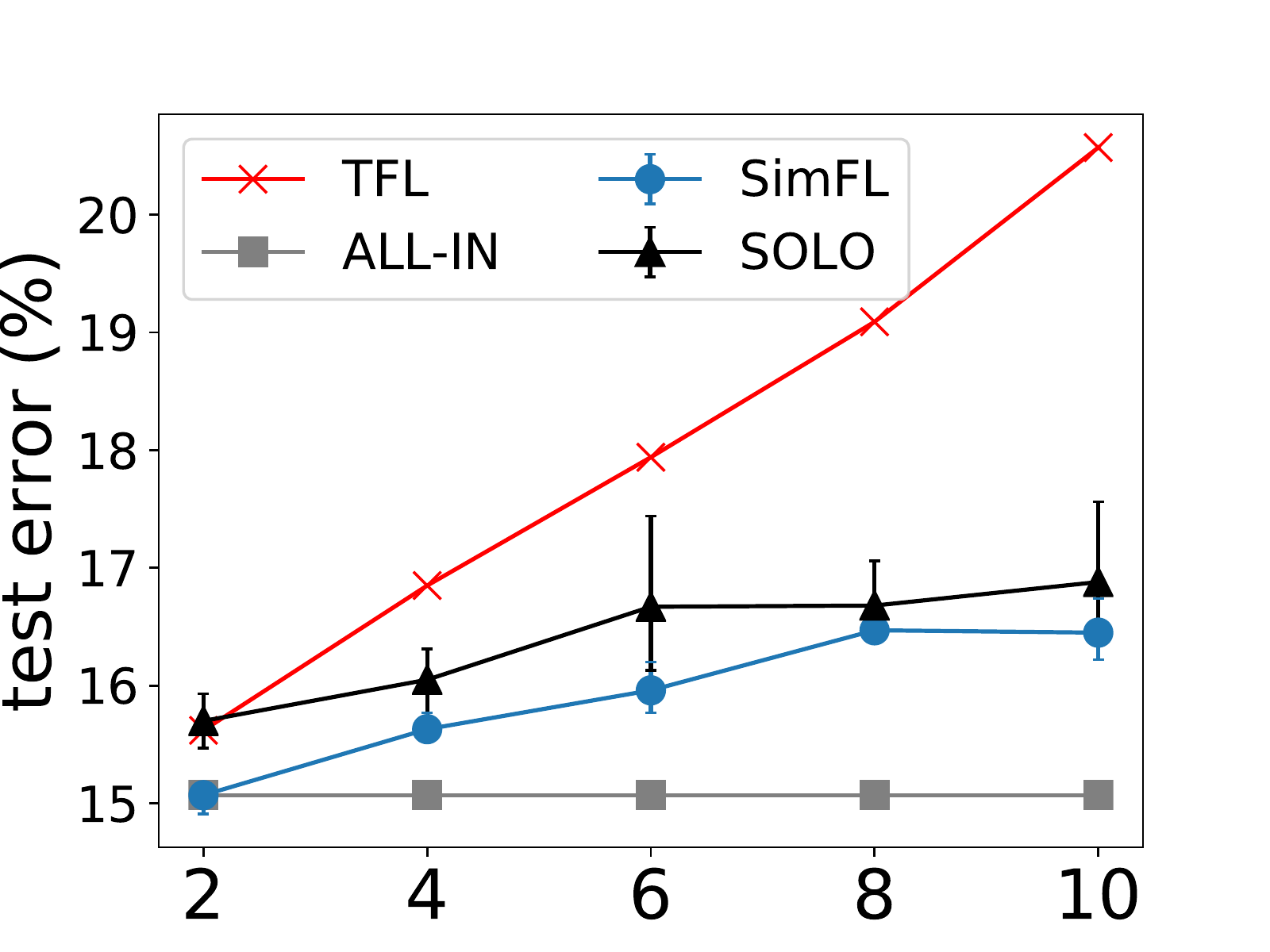}%
}
\subfloat[cod-rna]{\includegraphics[width=.475\columnwidth]{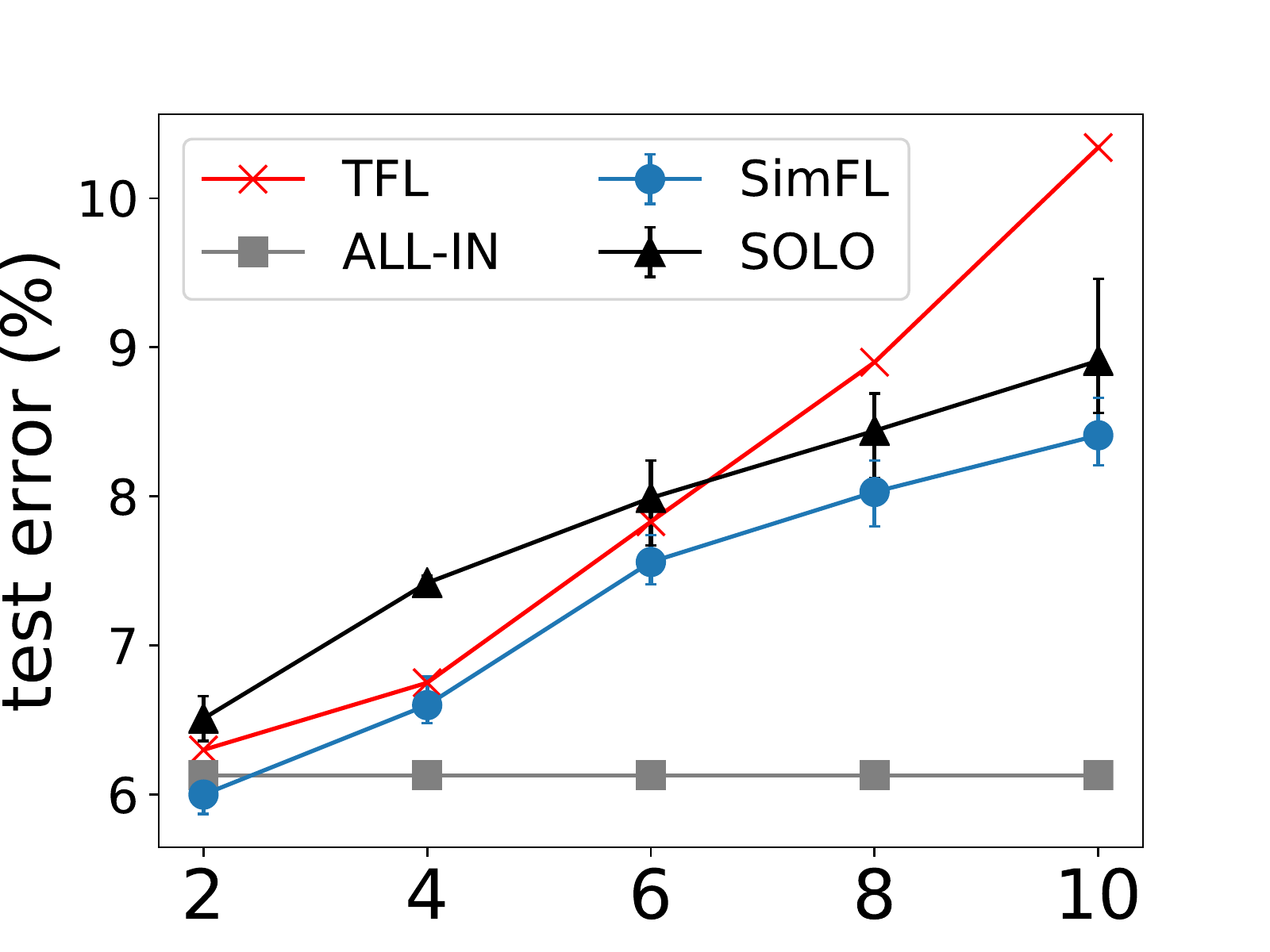}%
}
\hfil
\subfloat[real-sim]{\includegraphics[width=.475\columnwidth]{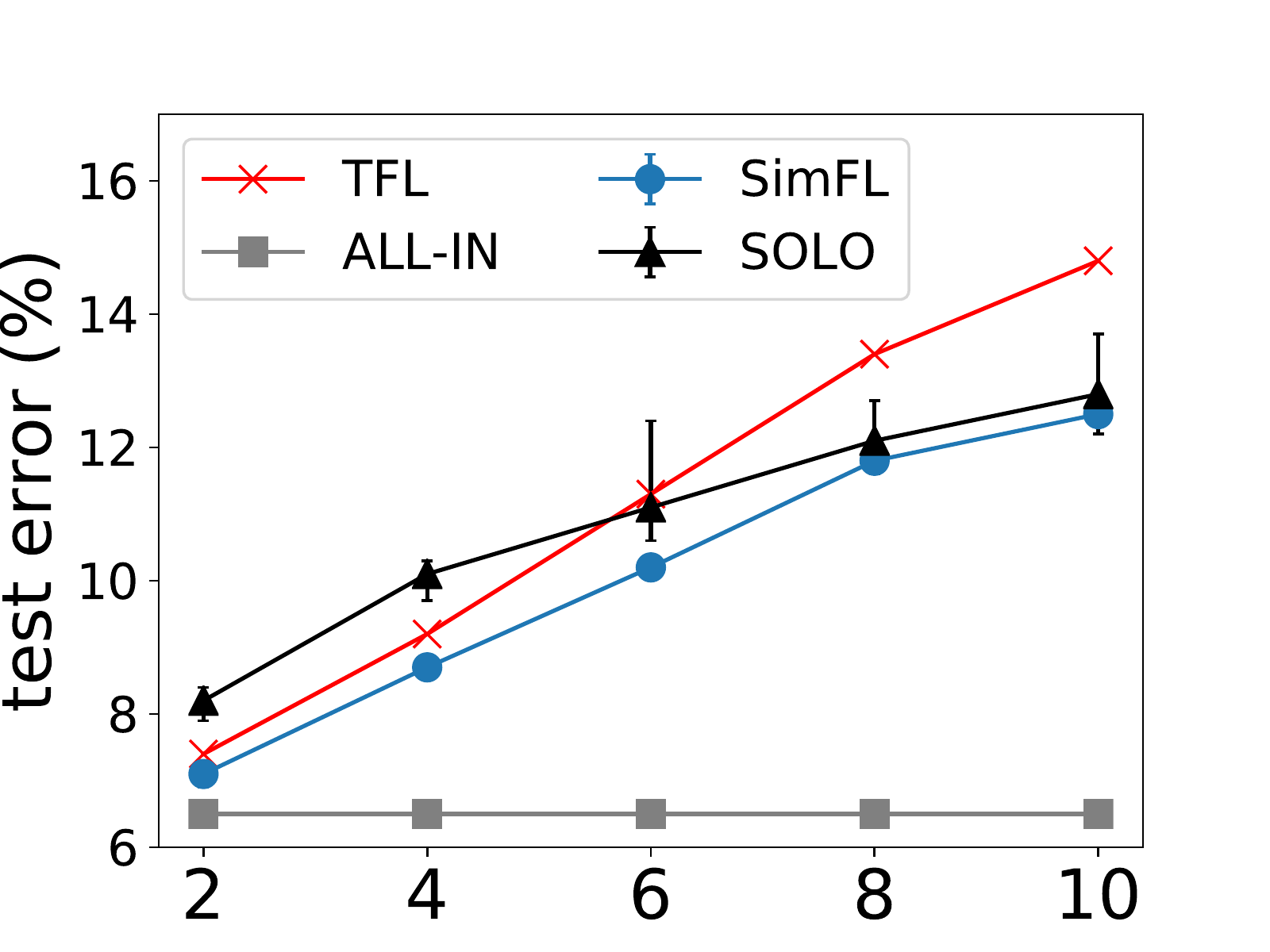}%
}
\subfloat[ijcnn1]{\includegraphics[width=.475\columnwidth]{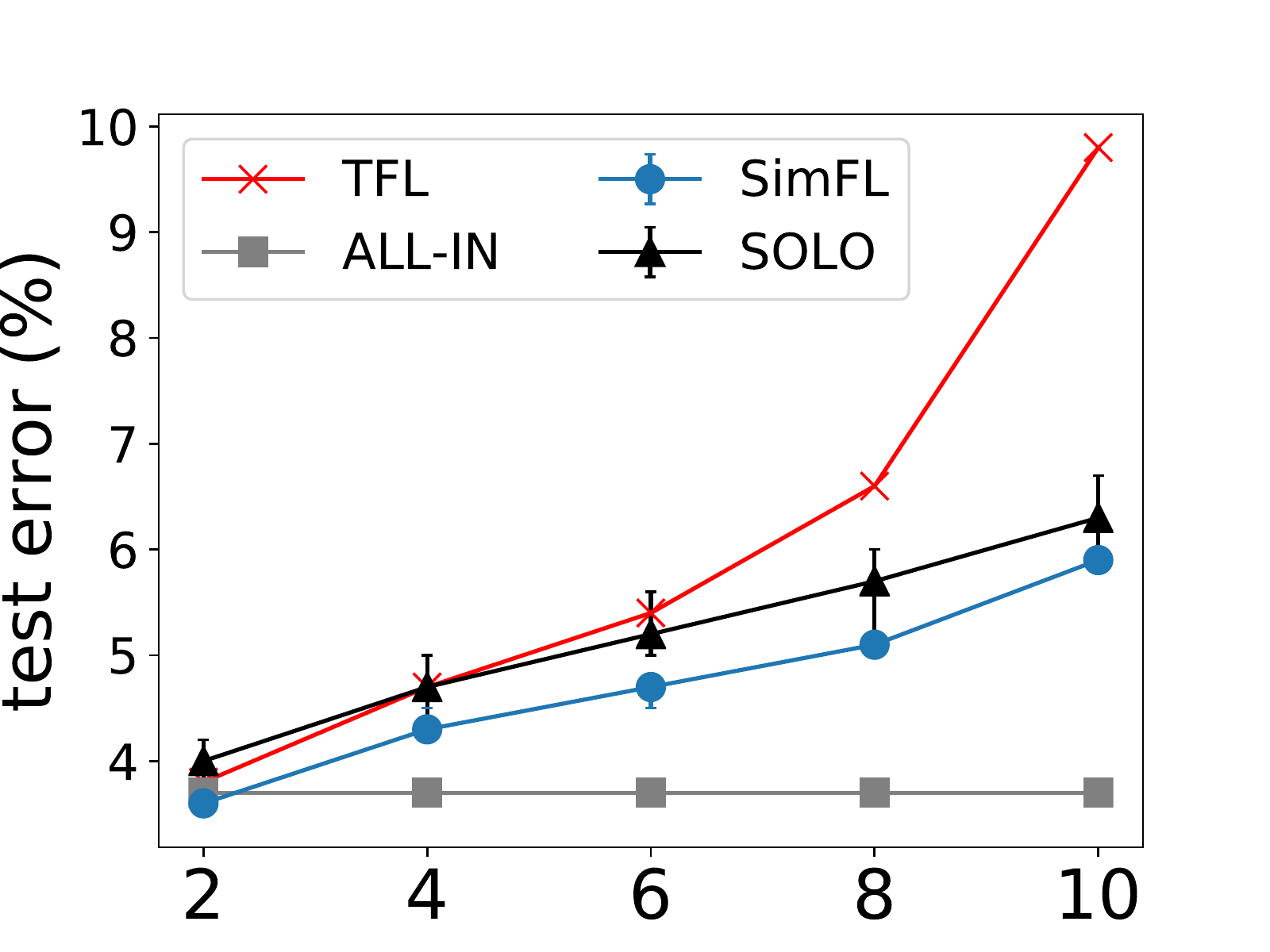}%
}
\hfil
\subfloat[SUSY]{\includegraphics[width=.475\columnwidth]{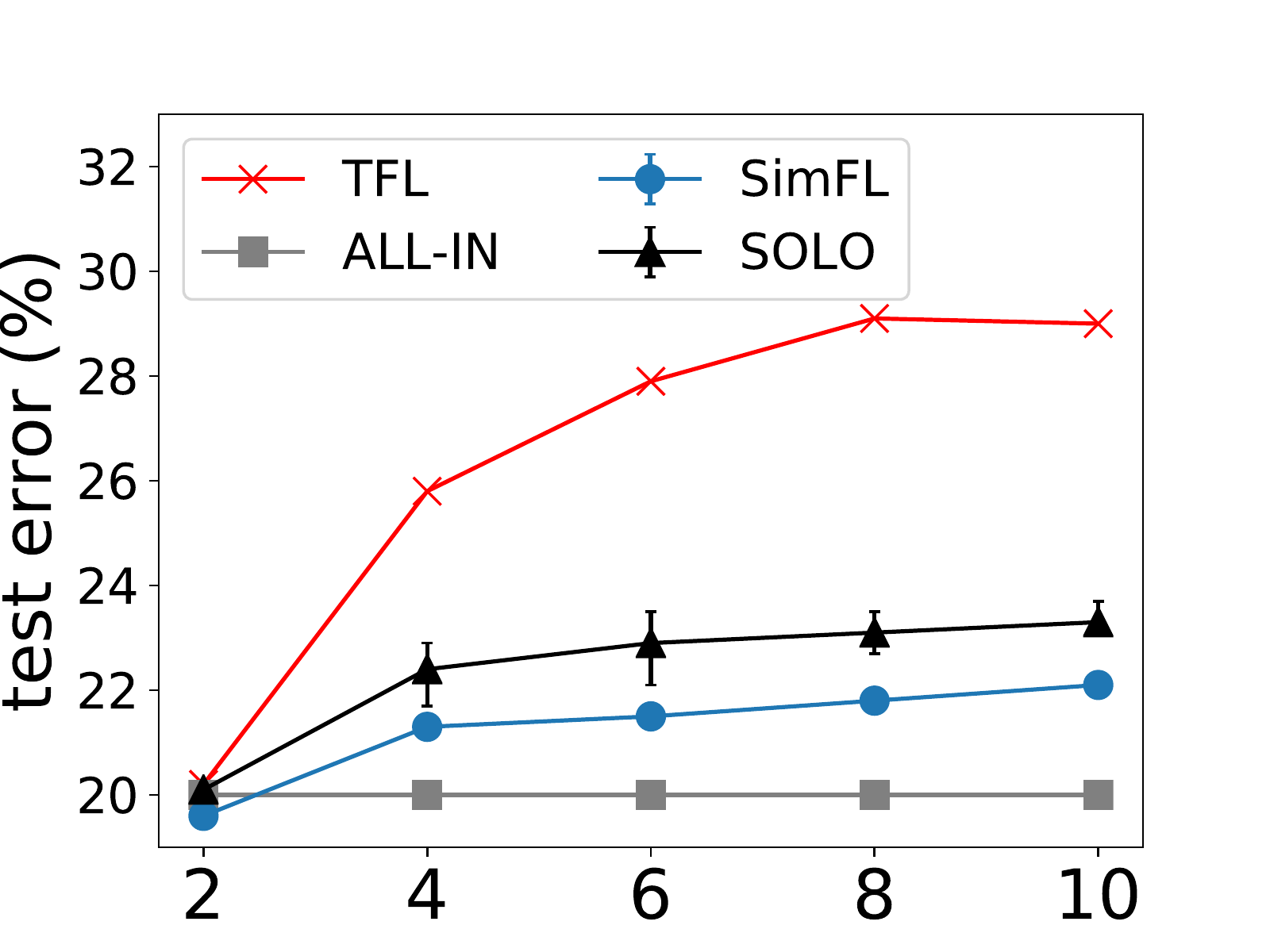}%
}
\subfloat[HIGGS]{\includegraphics[width=.475\columnwidth]{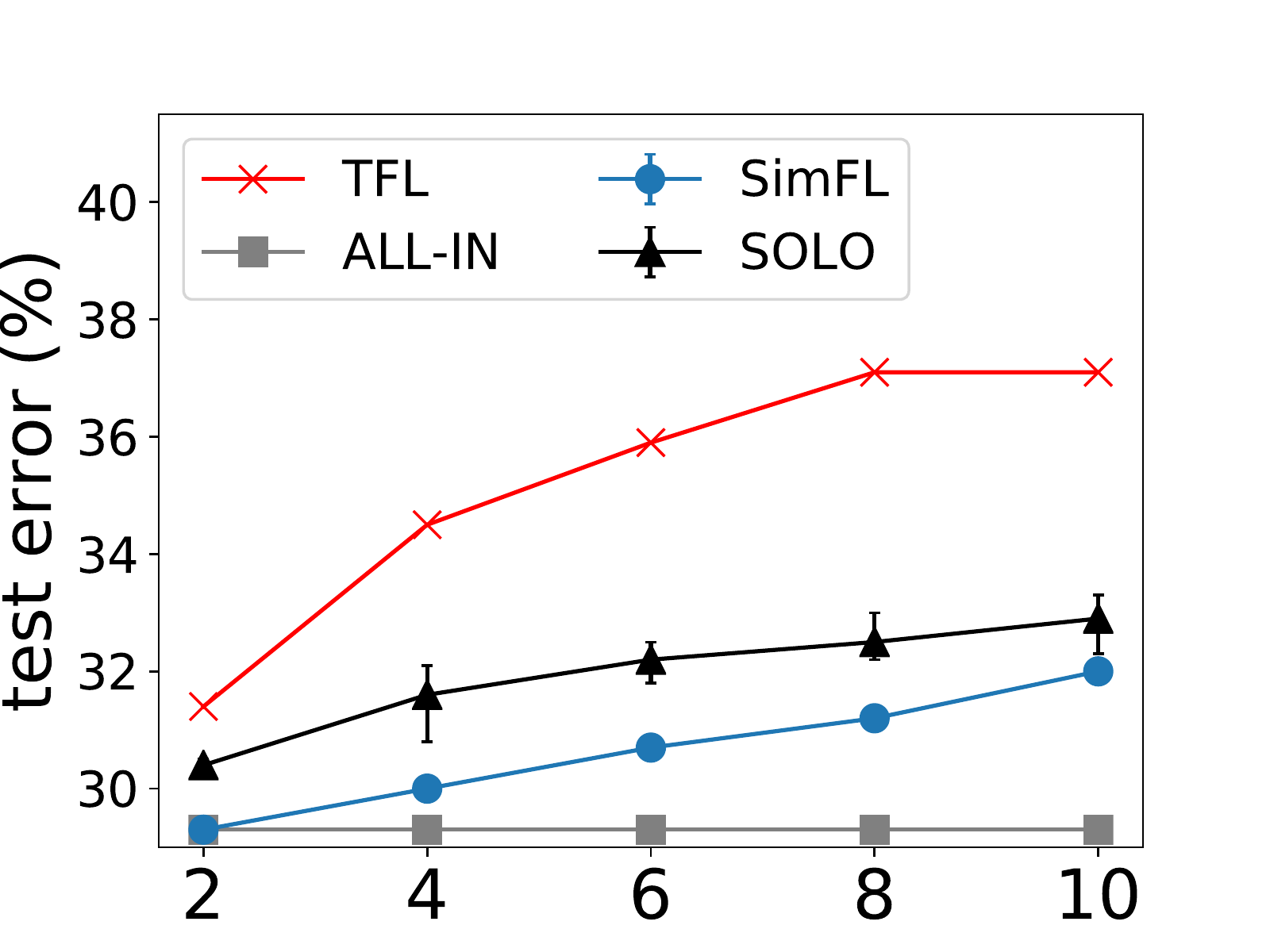}%
}
\caption{The impact of number of parties (balanced partition)}
\label{fig:errorbnp}
\end{figure}

Figure~\ref{fig:errorubnp} and Figure~\ref{fig:errorbnp} show the test errors with different number of parties. The ratio $\theta$ is set to $80\%$ for the unbalanced partition. As the number of parties increases, according to Section~\ref{sec:error_formula}, the upper bound of the generalization error of SimFL may increase since $N$ is fixed and $(N-N_m)$ increases, which agrees with the experimental results. Still, SimFL has a lower test error than the minimal error of SOLO at most times. While the test error of TFL changes dramatically as the number of parties increases, SimFL is much more stable in both balanced and unbalanced data partition.

\subsection{Efficiency}

To show the training time efficiency of SimFL, we present the time and communication cost in the training stage and the preprocessing stage (denoted as \emph{prep}) in Table~\ref{tbl:effi}. The number of parties is set to 10 and we adopt a balanced partition here. First, we can observe that the training time of SimFL is very close to SOLO. The computation overhead is less than 10\% of the total training time. Second, since SimFL only needs the local instances to build a tree while ALL-IN needs all the instances, the training time of SimFL is smaller than ALL-IN. Moreover, the proprecessing overhead is acceptable since it may only be required to perform once and reused for future training. One common scenario is that a user may try many runs of training with different hyper parameter settings and the pre-processing cost can be amortized among those runs. Last, the communication overhead per tree is very low, which costs no more than 10MB. In the encryption methods~\cite{liu2019boosting}, since additional keys need to be transferred, their communicated data size per tree can be much larger than ours. %Therefore, SimFL is an efficient and practical FL framework. 

\section{Conclusions}
The success of federated learning highly relies on the training time efficiency and the accuracy of the learned model. However, we find that existing horizontal federated learning systems of GBDTs suffer from low efficiency and/or low model accuracy. Based on an established relaxed privacy model, we propose a practical federated learning framework {\bf SimFL} for GBDTs by exploiting similarity. We take advantage of efficient locality-sensitive hashing to collect the similarity information without exposing the individual records, in contrast of costly secret sharing/encryption operations in previous studies. By designing a weighted gradient boosting method, we can utilize the similarity information to build decision trees with bounded errors. We prove that SimFL satisfies the privacy model. The experiments show that SimFL\del{ is an effective and efficient federated learning framework for GBDTs. It} significantly improves the predictive accuracy compared with training with data in individual parties alone, and is close to the model with joint data from all parties.

\begin{table}[h]
\centering
\caption{The training time (sec), preprocessing time (sec), communication cost (MB) per party in the preprocessing and communication cost (MB) per tree in the training.}
\label{tbl:effi}
\resizebox{.95\columnwidth}{!}{%
\begin{tabular}{|c|c|c|c|c|c|c|}
\hline
\multirow{3}{*}{datasets} & ALL-IN & SOLO & \multicolumn{4}{c|}{SimFL} \\ \cline{2-7} 
 & \multirow{2}{*}{\begin{tabular}[c]{@{}c@{}}training \\ time\end{tabular}} & \multirow{2}{*}{\begin{tabular}[c]{@{}c@{}}training\\  time\end{tabular}} & \multicolumn{2}{c|}{time} & \multicolumn{2}{c|}{communication} \\ \cline{4-7} 
 &  &  & training & prep & training & prep \\ \hline
a9a & 21.6 & 15.2 & 17.2 & 135 & 0.28 & 10.4 \\ \hline
cod-rna & 24.7 & 16.3 & 17.9 & 189 & 0.49 & 4.3 \\ \hline
real-sim & 69.5 & 32.4 & 34.5 & 380 & 0.6 & 23.1 \\ \hline
ijcnn1 & 29.3 & 15.7 & 17.4 & 228 & 0.42 & 8.4 \\ \hline
SUSY & 204.1 & 34.6 & 43.1 & 963 & 8.0 & 136 \\ \hline
HIGGS & 226.6 & 39.8 & 44.8 & 996 & 8.0 & 216 \\ \hline
\end{tabular}%
}
\end{table}

\section*{Acknowledgements}
This work is supported by a MoE AcRF Tier 1 grant (T1 251RES1824) and a MOE Tier 2 grant (MOE2017-T2-1-122) in Singapore. The authors thank Qiang Yang and Yang Liu for their insightful suggestions.
\bibliographystyle{aaai}
\bibliography{aaai20}

\begin{thebibliography}{}

\bibitem[\protect\citeauthoryear{Burges}{2010}]{burges2010ranknet}
Burges, C.~J.
\newblock 2010.
\newblock From ranknet to lambdarank to lambdamart: An overview.
\newblock {\em Learning} 11(23-581):81.

\bibitem[\protect\citeauthoryear{Chen and Guestrin}{2016}]{chen2016xgboost}
Chen, T., and Guestrin, C.
\newblock 2016.
\newblock Xgboost: A scalable tree boosting system.
\newblock In {\em Proceedings of the 22nd acm sigkdd international conference
  on knowledge discovery and data mining},  785--794.
\newblock ACM.

\bibitem[\protect\citeauthoryear{Cheng \bgroup et al\mbox.\egroup
  }{2019}]{cheng2019secureboost}
Cheng, K.; Fan, T.; Jin, Y.; Liu, Y.; Chen, T.; and Yang, Q.
\newblock 2019.
\newblock Secureboost: A lossless federated learning framework.
\newblock {\em arXiv preprint arXiv:1901.08755}.

\bibitem[\protect\citeauthoryear{Datar \bgroup et al\mbox.\egroup
  }{2004}]{datar2004locality}
Datar, M.; Immorlica, N.; Indyk, P.; and Mirrokni, V.~S.
\newblock 2004.
\newblock Locality-sensitive hashing scheme based on p-stable distributions.
\newblock In {\em Proceedings of the twentieth annual symposium on
  Computational geometry},  253--262.
\newblock ACM.

\bibitem[\protect\citeauthoryear{Du, Han, and Chen}{2004}]{du2004privacy}
Du, W.; Han, Y.~S.; and Chen, S.
\newblock 2004.
\newblock Privacy-preserving multivariate statistical analysis: Linear
  regression and classification.
\newblock In {\em SDM},  222--233.
\newblock SIAM.

\bibitem[\protect\citeauthoryear{Gionis \bgroup et al\mbox.\egroup
  }{1999}]{gionis1999similarity}
Gionis, A.; Indyk, P.; Motwani, R.; et~al.
\newblock 1999.
\newblock Similarity search in high dimensions via hashing.
\newblock In {\em Vldb}, volume~99,  518--529.

\bibitem[\protect\citeauthoryear{Ke \bgroup et al\mbox.\egroup
  }{2017}]{Ke2017LightGBMAH}
Ke, G.; Meng, Q.; Finley, T.; Wang, T.; Chen, W.; Ma, W.; Ye, Q.; and Liu,
  T.-Y.
\newblock 2017.
\newblock Lightgbm: A highly efficient gradient boosting decision tree.
\newblock In {\em NeurIPS}.

\bibitem[\protect\citeauthoryear{Kim \bgroup et al\mbox.\egroup
  }{2009}]{kim2009improving}
Kim, S.-M.; Pantel, P.; Duan, L.; and Gaffney, S.
\newblock 2009.
\newblock Improving web page classification by label-propagation over click
  graphs.
\newblock In {\em Proceedings of the 18th ACM conference on Information and
  knowledge management},  1077--1086.
\newblock ACM.

\bibitem[\protect\citeauthoryear{Ladyzhenskaia, Solonnikov, and
  Ural'ceva}{1968}]{ladyzhenskaia1968linear}
Ladyzhenskaia, O.~A.; Solonnikov, V.~A.; and Ural'ceva, N.~N.
\newblock 1968.
\newblock {\em Linear and quasi-linear equations of parabolic type}, volume~23.
\newblock American Mathematical Soc.

\bibitem[\protect\citeauthoryear{Li \bgroup et al\mbox.\egroup
  }{2019a}]{li2019federated}
Li, Q.; Wen, Z.; Wu, Z.; Hu, S.; Wang, N.; and He, B.
\newblock 2019a.
\newblock Federated learning systems: Vision, hype and reality for data privacy
  and protection.
\newblock {\em arXiv preprint arXiv:1907.09693}.

\bibitem[\protect\citeauthoryear{Li \bgroup et al\mbox.\egroup
  }{2019b}]{li2019privacy}
Li, Q.; Wu, Z.; Wen, Z.; and He, B.
\newblock 2019b.
\newblock Privacy-preserving gradient boosting decision trees.
\newblock {\em AAAI-20, arXiv preprint arXiv:1911.04209}.

\bibitem[\protect\citeauthoryear{Liu \bgroup et al\mbox.\egroup
  }{2019}]{liu2019boosting}
Liu, Y.; Ma, Z.; Liu, X.; Ma, S.; Nepal, S.; and Deng, R.
\newblock 2019.
\newblock Boosting privately: Privacy-preserving federated extreme boosting for
  mobile crowdsensing.
\newblock {\em arXiv preprint arXiv:1907.10218}.

\bibitem[\protect\citeauthoryear{Liu, Chen, and Yang}{2018}]{liu2018secure}
Liu, Y.; Chen, T.; and Yang, Q.
\newblock 2018.
\newblock Secure federated transfer learning.
\newblock {\em arXiv preprint arXiv:1812.03337}.

\bibitem[\protect\citeauthoryear{McMahan \bgroup et al\mbox.\egroup
  }{2016}]{McMahan2016FederatedLO}
McMahan, H.~B.; Moore, E.; Ramage, D.; and y~Arcas, B.~A.
\newblock 2016.
\newblock Federated learning of deep networks using model averaging.
\newblock {\em CoRR} abs/1602.05629.

\bibitem[\protect\citeauthoryear{Mirhoseini, Sadeghi, and
  Koushanfar}{2016}]{mirhoseini2016cryptoml}
Mirhoseini, A.; Sadeghi, A.-R.; and Koushanfar, F.
\newblock 2016.
\newblock Cryptoml: Secure outsourcing of big data machine learning
  applications.
\newblock In {\em 2016 IEEE International Symposium on Hardware Oriented
  Security and Trust (HOST)}.

\bibitem[\protect\citeauthoryear{Mohri, Sivek, and
  Suresh}{2019}]{pmlr-v97-mohri19a}
Mohri, M.; Sivek, G.; and Suresh, A.~T.
\newblock 2019.
\newblock Agnostic federated learning.
\newblock In Chaudhuri, K., and Salakhutdinov, R., eds., {\em ICML}, volume~97
  of {\em Proceedings of Machine Learning Research},  4615--4625.
\newblock PMLR.

\bibitem[\protect\citeauthoryear{Patarasuk and
  Yuan}{2009}]{patarasuk2009bandwidth}
Patarasuk, P., and Yuan, X.
\newblock 2009.
\newblock Bandwidth optimal all-reduce algorithms for clusters of workstations.
\newblock {\em Journal of Parallel and Distributed Computing} 69(2):117--124.

\bibitem[\protect\citeauthoryear{Qi \bgroup et al\mbox.\egroup
  }{2017}]{qi2017distributed}
Qi, L.; Zhang, X.; Dou, W.; and Ni, Q.
\newblock 2017.
\newblock A distributed locality-sensitive hashing-based approach for cloud
  service recommendation from multi-source data.
\newblock {\em IEEE Journal on Selected Areas in Communications}
  35(11):2616--2624.

\bibitem[\protect\citeauthoryear{Richardson, Dominowska, and
  Ragno}{2007}]{richardson2007predicting}
Richardson, M.; Dominowska, E.; and Ragno, R.
\newblock 2007.
\newblock Predicting clicks: estimating the click-through rate for new ads.
\newblock In {\em Proceedings of the 16th international conference on World
  Wide Web},  521--530.
\newblock ACM.

\bibitem[\protect\citeauthoryear{Shalev-Shwartz and
  Ben-David}{2014}]{shalev2014understanding}
Shalev-Shwartz, S., and Ben-David, S.
\newblock 2014.
\newblock {\em Understanding machine learning: From theory to algorithms}.
\newblock Cambridge university press.

\bibitem[\protect\citeauthoryear{Shi \bgroup et al\mbox.\egroup
  }{2017}]{shi2017distributed}
Shi, E.; Chan, T.-H.~H.; Rieffel, E.; and Song, D.
\newblock 2017.
\newblock Distributed private data analysis: Lower bounds and practical
  constructions.
\newblock {\em ACM Transactions on Algorithms (TALG)} 13(4):50.

\bibitem[\protect\citeauthoryear{Smith \bgroup et al\mbox.\egroup
  }{2017}]{smith2017federated}
Smith, V.; Chiang, C.-K.; Sanjabi, M.; and Talwalkar, A.~S.
\newblock 2017.
\newblock Federated multi-task learning.
\newblock In {\em Advances in Neural Information Processing Systems},
  4424--4434.

\bibitem[\protect\citeauthoryear{Takabi, Hesamifard, and
  Ghasemi}{2016}]{takabi2016privacy}
Takabi, H.; Hesamifard, E.; and Ghasemi, M.
\newblock 2016.
\newblock Privacy preserving multi-party machine learning with homomorphic
  encryption.
\newblock In {\em 29th Annual Conference on Neural Information Processing
  Systems}.

\bibitem[\protect\citeauthoryear{Wang \bgroup et al\mbox.\egroup
  }{2014}]{wang2014privacy}
Wang, B.; Yu, S.; Lou, W.; and Hou, Y.~T.
\newblock 2014.
\newblock Privacy-preserving multi-keyword fuzzy search over encrypted data in
  the cloud.
\newblock In {\em IEEE INFOCOM 2014-IEEE Conference on Computer
  Communications},  2112--2120.
\newblock IEEE.

\bibitem[\protect\citeauthoryear{Wen \bgroup et al\mbox.\egroup
  }{2019}]{ThunderGBM}
Wen, Z.; Shi, J.; Li, Q.; He, B.; and Chen, J.
\newblock 2019.
\newblock Thundergbm: Fast gbdts and random forests on gpus.
\newblock In {\em https://github.com/Xtra-Computing/thundergbm}.

\bibitem[\protect\citeauthoryear{Yang \bgroup et al\mbox.\egroup
  }{2019}]{Yang:2019:FML:3306498.3298981}
Yang, Q.; Liu, Y.; Chen, T.; and Tong, Y.
\newblock 2019.
\newblock Federated machine learning: Concept and applications.
\newblock {\em ACM Trans. Intell. Syst. Technol.} 10(2):12:1--12:19.

\bibitem[\protect\citeauthoryear{Yao}{1982}]{yao1982protocols}
Yao, A. C.-C.
\newblock 1982.
\newblock Protocols for secure computations.
\newblock In {\em FOCS}, volume~82,  160--164.

\bibitem[\protect\citeauthoryear{Yurochkin \bgroup et al\mbox.\egroup
  }{2019}]{pmlr-v97-yurochkin19a}
Yurochkin, M.; Agarwal, M.; Ghosh, S.; Greenewald, K.; Hoang, N.; and Khazaeni,
  Y.
\newblock 2019.
\newblock {B}ayesian nonparametric federated learning of neural networks.
\newblock In Chaudhuri, K., and Salakhutdinov, R., eds., {\em ICML}, volume~97
  of {\em Proceedings of Machine Learning Research},  7252--7261.
\newblock PMLR.

\bibitem[\protect\citeauthoryear{Zhao \bgroup et al\mbox.\egroup
  }{2018}]{zhao2018inprivate}
Zhao, L.; Ni, L.; Hu, S.; Chen, Y.; Zhou, P.; Xiao, F.; and Wu, L.
\newblock 2018.
\newblock Inprivate digging: Enabling tree-based distributed data mining with
  differential privacy.
\newblock In {\em INFOCOM},  2087--2095.
\newblock IEEE.

\bibitem[\protect\citeauthoryear{Zolotarev}{1986}]{zolotarev1986one}
Zolotarev, V.~M.
\newblock 1986.
\newblock {\em One-dimensional stable distributions}, volume~65.
\newblock American Mathematical Soc.

\end{thebibliography}

\appendix
\section{Proof of Theorem 1}
\begin{proof}
Without loss of generality, we only need to prove that the protocol is secure against $P_m$. In the whole FL process, $P_m$ knows the hash values of the other instances and the aggregated gradients of the similar instances. Next, we prove that there are infinite number of instances that can provide the same hash values and gradients.

Given the hash values, $P_m$ knows $L$ different compound inequalities about an instance $\mathbf{x}_k$ with the form $\mathcal{F}_j(\mathbf{x}_k) \leq \frac{\mathbf{a}_j\cdot\mathbf{x}_k+b_j}{r} < \mathcal{F}_j(\mathbf{x}_k) + 1$, $j=1..L$. Consider a stricter system of $L$ different linear equations $\frac{\mathbf{a}_j\cdot\mathbf{x}_k+b_j}{r} = z_j$, where $z_j\in [\mathcal{F}_j(\mathbf{x}_k),\mathcal{F}_j(\mathbf{x}_k) + 1)$. Obviously, for a linear system, there are no solution or an infinite number of solutions if the number of unknowns (i.e., $d$) is bigger than the number of equations (i.e., $L$)~\cite{ladyzhenskaia1968linear}. Since we already know there is at least one solution (i.e., $\mathbf{x}_k$), the number of solutions is infinite. Thus, there are infinite number of instances that can result in the same hash values that $P_m$ knows. Since the gradients are computed based on the prediction values, the instances can provide the same gradient as long as they have the same prediction values in the trees. Note that the restrictions in the decision trees only specify a range of the feature values. So we have infinite number of instances that can provide the same gradients as long as they satisfy the same restrictions on the features values and are divided into the same leaf. Thus, we can have infinite number of instances that can provide the same output for party $P_m$.
\end{proof}

\section{Proof of Theorem 2}
\begin{proof}
    For ease of presentation, we suppose the instances have different globally IDs (i.e., for any two instances $\mathbf{x}_i^m$ and $\mathbf{x}_j^n$, if $m\neq n$ then $i\neq j$). Thus, we can use $g_i$ instead of $g_i^m$ to denote the first order gradient of an instance $\mathbf{x}_i^m$. Also, we use $h_i$ instead of $h_i^m$ to denote the second order gradient of $\mathbf{x}_i^m$. We have
\begin{equation}
\begin{aligned}
    \mathcal{\tilde{L}}^{(t)} &= \sum_{i=1}^{n} [g_i f_t(\mathbf{x}_i)+\frac{1}{2}h_i f_t^2(\mathbf{x}_i)]+\Omega(f_t)\\
    &= \sum_{\mathbf{x}_q^m\in I_m}[g_q f_t(\mathbf{x}_q^m)+\frac{1}{2} h_q f_t^2(\mathbf{x}_q^m)] \\
    & +\sum_{\substack{j=1\\j\neq m}}^M\sum_{\mathbf{x}_r^j \in I_j}[g_r f_t(\mathbf{x}_r^j)+\frac{1}{2} h_r f_t^2(\mathbf{x}_r^j)] + \Omega(f_t)\\
\end{aligned}
\end{equation}  
With Eq.~\eqref{eq:app_loss}, we have
\begin{equation}
\begin{aligned}
     \mathcal{\tilde{L}}^{(t)}&=\mathcal{\tilde{L}}_w^{(t)} - \sum_{\substack{j=1\\j\neq m}}^M\sum_{\substack{r\\\mathbf{x}_r^j\in I_j}}[g_r f_t(\mathbf{x}_{\mtx{S}_{rm}^j}^m) +\frac{1}{2}h_rf_t^2(\mathbf{x}_{\mtx{S}_{rm}^j}^m)]  \\
    &  +\sum_{\substack{j=1\\j\neq m}}^M\sum_{\mathbf{x}_r^j \in I_j}[g_r f_t(\mathbf{x}_r^j)+\frac{1}{2} h_r f_t^2(\mathbf{x}_r^j)]\\
    &=\mathcal{\tilde{L}}_w^{(t)} - \sum_{\substack{j=1\\j\neq m}}^M\sum_{\substack{r\\\mathbf{x}_r^j\in I_j}}[g_r (f_t(\mathbf{x}_{\mtx{S}_{rm}^j}^m) - f_t(\mathbf{x}_r^j)) \\
    &+\frac{1}{2}h_r(f_t^2(\mathbf{x}_{\mtx{S}_{rm}^j}^m)-f_t^2(\mathbf{x}_r^j))]
\end{aligned}
\end{equation}
Thus, we have
\begin{equation}
\begin{aligned}
    \varepsilon^t &= |\mathcal{\tilde{L}}_w^{(t)} - \mathcal{\tilde{L}}^{(t)}| \\
    &= \sum_{\substack{j=1\\j\neq m}}^M\sum_{\substack{r\\\mathbf{x}_r^j\in I_j}}|g_r (f_t(\mathbf{x}_{\mtx{S}_{rm}^j}^m) - f_t(\mathbf{x}_r^j))\\
    &  +\frac{1}{2}h_r(f_t^2(\mathbf{x}_{\mtx{S}_{rm}^j}^m)-f_t^2(\mathbf{x}_r^j))|
    % \varepsilon^t =  \sum_{x_r \notin I_m}|[g_r (f_t(x_r^m) - f_t(x_r))+\frac{1}{2}h_r(f_t^2(x_r^m)-f_t^2(x_r))]|
\end{aligned}
\end{equation}
Let $\xi_r = |g_r (f_t(\mathbf{x}_{\mtx{S}_{rm}^j}^m) - f_t(\mathbf{x}_r^j))+\frac{1}{2}h_r(f_t^2(\mathbf{x}_{\mtx{S}_{rm}^j}^m)-f_t^2(\mathbf{x}_r^j))|$. Since $g'=\max_i |g_i|$, $h'=\max_i |h_i|$, and $f_t'=\max_i |f_t(\mathbf{x}_i)|$, we have

\begin{equation}
\label{eq:xi_bound}
    \xi_r \leq 2g' f_t'+\frac{1}{2}h' f_t'^2
\end{equation}

Notice that $\xi_r = 0$ if $\mathbf{x}_{\mtx{S}_{rm}^j}^m$ and $\mathbf{x}_r^j$ go to the same leaf of the tree $f_t$. According to our assumption, the feature values of each feature are i.i.d. uniform random variables and the split value is randomly chosen from the feature values. Let $d_t = \max_{r,j}\norm{\mathbf{x}_{\mtx{S}_{rm}^j}^m-\mathbf{x}_r^j}_1$ and $d_{m} = \max_{i,j}\norm{\mathbf{x}_i- \mathbf{x}_j}_1$, then the probability that $\mathbf{x}_{\mtx{S}_{rm}^j}^m$ and $\mathbf{x}_r^j$ are divided into two directions in each node is smaller than $\frac{d_t}{d_{m}}$ (i.e., the probability that randomly drop a ball in a line with length $d_m$ and it falls in a interval with length $d_t$). Let $D$ denotes the depth of the tree. Then, the probability that $\mathbf{x}_{\mtx{S}_{rm}^j}^m$ and $\mathbf{x}_r^j$ go to the same leaf is bigger than $(1-\frac{d_t}{d_{m}})^D$. There are $(N-N_m)$ different $\xi_r$. Let $H$ denotes the nubmer of times that $\xi_r\neq0$. By Hoeffding's inequality with a Bernoulli distribution, with probability at least $1-\delta$, we have
\begin{equation}
\label{eq:h_bound}
    H \leq [1-(1-\frac{d_t}{d_m})^D](N-N_m)+\sqrt{\frac{(N-N_m)\ln{\frac{1}{\delta}}}{2}}
\end{equation}
With Eq.~\eqref{eq:xi_bound} and Eq.~\eqref{eq:h_bound}, we have
\begin{equation}
\begin{aligned}
    \label{eq:errbound}
    \varepsilon^t \leq &\Big([1-(1-\frac{d_t}{d_m})^D](N-N_m)+\sqrt{\frac{(N-N_m)\ln{\frac{1}{\delta}}}{2}}\Big) \\
    &\cdot (2g' f_t'+\frac{1}{2}h' f_t'^2)
\end{aligned}
\end{equation}

\end{proof}
\end{document}